\documentclass[accepted]{uai2026} 
                        
\usepackage[american]{babel}

\usepackage{natbib} 
\usepackage{mathtools} 
\usepackage{booktabs} 
\usepackage{tikz} 
\usepackage{amsmath, amsthm, amsfonts}
\usepackage[ruled, vlined, boxed, linesnumbered]{algorithm2e}
\usepackage{threeparttable}
\usepackage{multirow}
\usepackage{subcaption}
\usepackage{tabularx}
\usepackage{booktabs} 
\usepackage[most]{tcolorbox}
\usepackage{xcolor}
\usepackage{amssymb} 
\usepackage{pifont}  \usepackage{titletoc}

\newtheorem{definition}{Definition}
\newtheorem{theorem}{Theorem}
\newtheorem*{theorem*}{Theorem}

\newtcolorbox{quespromptbox}[1]{
    colback=gray!5!white,
    colframe=gray!75!black, 
    fonttitle=\bfseries,
    title=#1,              
    enhanced,
    attach boxed title to top left={yshift=-2mm, xshift=2mm},
    boxed title style={colback=gray!75!black},
    rounded corners=all,
    drop shadow,
    before skip=10pt,
    after skip=10pt
}
\newtcolorbox{longformbox}[2]{
    colback=white,
    colframe=blue!50!black,
    fonttitle=\bfseries,
    title={Dataset: #1, \hfill Model: #2},
    enhanced,
    breakable, 
    attach boxed title to top left={yshift=-2mm, xshift=2mm},
    boxed title style={colback=blue!50!black},
    rounded corners=all,
    fontupper=\small,
    before skip=15pt,
    after skip=15pt,
    middle=1mm,
    segmentation style={solid, blue!50!black}
}
\newcommand{\boxsection}[1]{\noindent\textbf{\textcolor{blue!50!black}{#1}}}

\newcommand{\correct}{\textcolor{green!70!black}{\ding{51}}} 
\newcommand{\wrong}{\textcolor{red}{\ding{55}}}             

\title{SeSE: Black-Box Uncertainty Quantification for \\Large Language Models Based on Structural Information Theory}

\author[1]{Xingtao Zhao}
\author[1]{Hao Peng\thanks{Corresponding author.}}
\author[2]{Dingli Su}
\author[2]{Xianghua Zeng}
\author[3]{Chunyang Liu}
\author[4]{Jinzhi Liao}
\author[5]{Philip S. Yu}
\affil[1]{%
    School of Cyber Science and Technology\\
    Beihang University\\
    Beijing, China
}
\affil[2]{%
    School of Computer Science and Engineering\\
    Beihang University\\
    Beijing, China
}
\affil[3]{%
    Didi Chuxing\\
    Beijing, China
  }
\affil[4]{%
    Laboratory for Big Data and Decision\\
    National University of Defense Technology\\ Changsha, China
  }
\affil[5]{%
    Department of Computer Science\\
    University of Illinois Chicago\\ 
    Chicago, USA
}
  
\vspace{1em}
\begin{document}
\maketitle

\begin{abstract}
Reliable uncertainty quantification (UQ) is essential for deploying large language models (LLMs) in safety-critical scenarios, as it enables them to abstain from responding when uncertain, thereby avoiding hallucinations, i.e., plausible yet factually incorrect responses. However, while current semantic UQ methods have achieved state-of-the-art performance, they inherently overlook latent semantic structural information that could enable more precise uncertainty estimates. In this paper, we propose \underline{Se}mantic \underline{S}tructural \underline{E}ntropy ({SeSE}), a principled black-box UQ framework applicable to both open- and closed-source LLMs. 
To reveal the intrinsic structure of the LLM semantic space, SeSE constructs its \emph{hierarchical abstraction} based on the principle of structural entropy minimization. The structural entropy of the resulting optimal hierarchical abstraction thus quantifies the inherent uncertainty within the semantic space after optimal compression. 
Additionally, unlike existing methods that primarily focus on simple short-form generation, we extend SeSE to provide interpretable and granular uncertainty estimation for long-form outputs.
We theoretically prove that SeSE generalizes semantic entropy, the gold standard for UQ in LLMs, and empirically demonstrate its superior performance over baselines across 24 model-dataset combinations.
\end{abstract}

\section{Introduction}\label{sec:intro}
Large language models (LLMs) have been widely adopted across various fields \citep{perplexity2025,11005661}, owing to their impressive general intelligence. However, even state-of-the-art (SOTA) models frequently generate plausible yet incorrect statements \citep{openai2025gpt5}, a phenomenon known as \emph{hallucination} \citep{huang2025survey}, which impedes their deployment in risk-sensitive domains. Despite extensive research seeking to eliminate hallucinations \citep{rafailov2023direct,liu2024decodingtime}, complete solutions remain elusive. A promising approach for avoiding hallucinations is uncertainty quantification (UQ) \citep{lingenerating}, which estimates the likelihood of an LLM hallucinating falsehoods for a given input \citep{cole2023selectively}. Lower uncertainty suggests the initial response is acceptable, whereas higher uncertainty should trigger LLMs to abstain from answering and alert users to potential errors.

However, the open-ended nature of LLM generation hinders the direct application of traditional UQ methods \citep{liu2020simple,malinin2021uncertaintyestimationautoregressivestructured}, which treat LLM outputs as autoregressive sequences and consider only lexical uncertainty. Since response correctness fundamentally depends on semantics, and distinct token sequences can convey identical meanings, uncertainty in the semantic space serves as a more reliable indicator of trustworthiness than lexical uncertainty. To address this issue, Semantic Entropy (SE) \citep{farquhar2024detecting} was proposed to quantify uncertainty at the semantic rather than token level, serving as the gold standard for UQ in LLMs \citep{huang2025survey}.

\begin{figure*}[!ht]
\centering
\includegraphics[width=1\linewidth,trim={0cm 0 0cm 0}, clip]{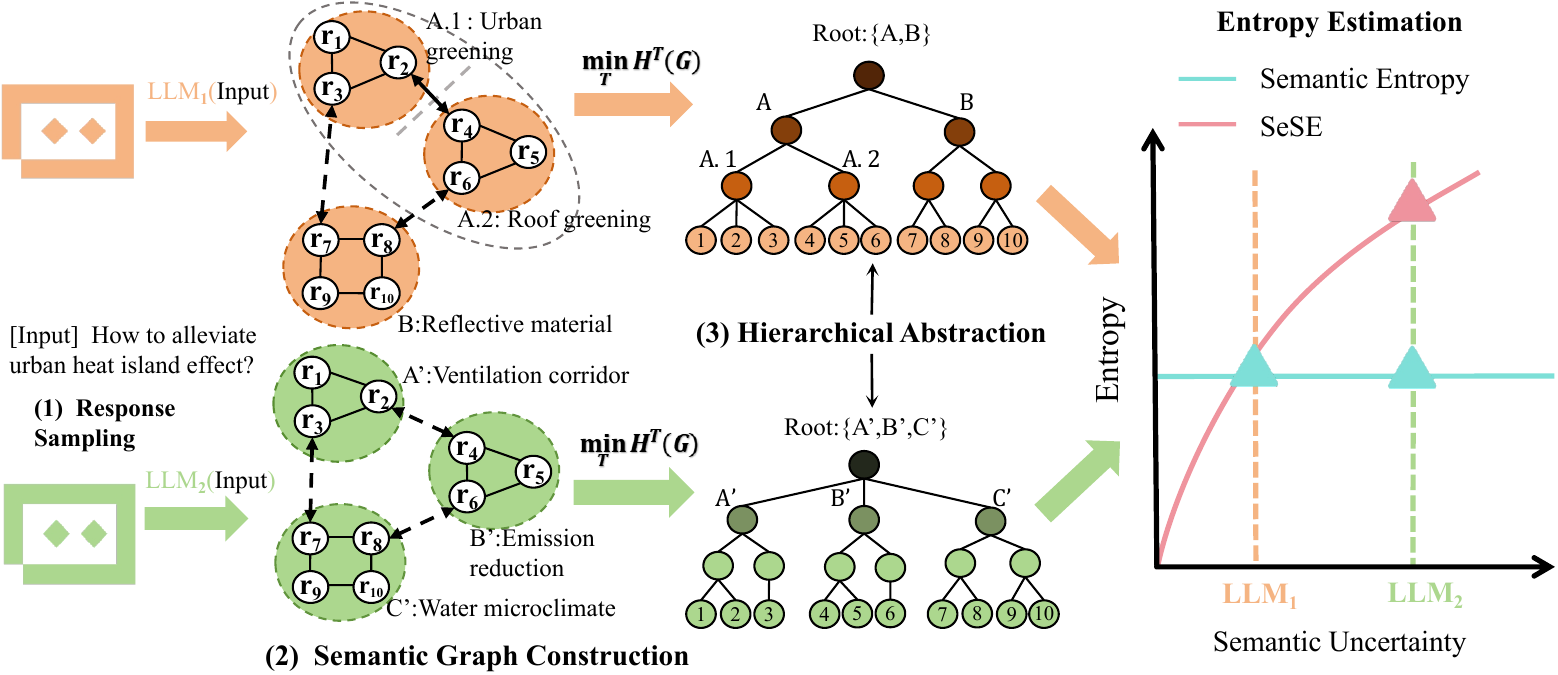}

\caption{Illustration of SeSE in short-form generation. 
Superficially, both semantic spaces in part 2 appear similar, as they each contain three semantic substructures. 
However, the hierarchical abstraction (part 3) of $\text{LLM}_1$ is notably more regular, featuring a hierarchically nested structure where the high-level substructure A aggregates two finer-grained substructures, facilitating efficient compression. In contrast, the hierarchical abstraction of $\text{LLM}_2$ is more disordered and resists compression. This implies $\text{LLM}_1$ exhibits lower uncertainty than $\text{LLM}_2$, i.e., $\text{LLM}_1$ is more certain that "greening" is an excellent solution.
SeSE captures the inherent semantic structure by constructing the optimal hierarchical abstraction, which is formalized by an encoding tree with minimal structural entropy, thereby correctly assigning $\text{LLM}_1$ a lower uncertainty score.
}
\label{fig:SeSE-sentence}
\end{figure*}

Despite its success, SE and its subsequent extensions \citep{nikitin2024kernel,QiuM24,LiSYJCCR25} suffer from key limitations. First, they fail to capture the inherent semantic structure that defines the organizational principle of LLM semantic space \citep{li2024survey}. As illustrated in part 2 of Figure~\ref{fig:SeSE-sentence}, semantic spaces are often hierarchically organized, with substructures recursively containing sub-substructures. According to the ``Compositional Similarity'' principle \citep{boiman2006similarity}, identifying this hierarchical substructures of semantic spaces can contribute to more precise uncertainty estimates, as differences between substructures could help distinguish superficially similar semantic spaces.
Second, while current semantic UQ methods have progressed in short-form settings where generations typically consist of one or two sentences \citep{farquhar2024detecting,nikitin2024kernel,QiuM24,LiSYJCCR25}, they often lack sufficient granularity for real-world applications in which LLMs often generate long-form paragraphs comprising interwoven true and false claims. Recent works \citep{manakul2023selfcheckgpt,mohri2024language,zhang-etal-2024-luq,jiang2024graph} have begun to quantify claim-level uncertainty based on the concept of self-consistency \citep{manakul2023selfcheckgpt}. However, they primarily rely on heuristic sample-and-count techniques, which offer limited interpretability and fail to capture fine-grained semantic dependencies between claims and responses.

To address these issues, we propose {Semantic Structural Entropy} (SeSE), a principled black-box UQ framework that does not require access to LLM internal states and is applicable to both open- and closed-source LLMs. 
For the first issue, to represent the intrinsic hierarchical structure of the LLM semantic space, SeSE constructs its optimal \emph{hierarchical abstraction} adhering to the structural entropy minimization principle \citep{li2016structural}, which is widely used to discover the natural hierarchical structure of graph data. The structural entropy of this hierarchical abstraction therefore quantifies the inherent semantic uncertainty of LLMs.
For the second issue, we extend SeSE to provide interpretable and granular claim-level uncertainty estimation in long-form generation. Following prior decomposition methods \citep{min2023factscore}, we segment long-form outputs into atomic claims and construct a claim-response bipartite graph to capture fine-grained semantic dependencies. The SeSE of a claim is defined as the uncertainty of reaching that claim via random walks on this graph.
Importantly, we theoretically prove that SeSE can recover SE when the encoding tree is restricted to a single layer.
In summary, the contributions of this paper are\footnote{The code and data is available at: \url{https://github.com/SELGroup/SeSE}}: 
\begin{itemize}[leftmargin=\parindent, labelindent=0pt]
\item We propose SeSE, a principled black-box UQ framework built upon structural information theory and applicable to both open- and closed-source LLMs. To the best of our knowledge, this is the \emph{first} work to incorporate semantic structural information into UQ for LLMs.
\item We extend SeSE to provide interpretable, granular claim-level UQ in long-form generation by modeling random semantic interactions within claim-response bipartite graphs.
\item We prove that SeSE is a generalization of semantic entropy.
\item We empirically compare SeSE with baselines methods across 24 model-dataset pairs, achieving SOTA results.
\end{itemize}

\section{Preliminaries}
\label{sec:Pre}

\paragraph{Problem Formulation}
Given a pre-trained LLM $\mathcal{M}$, SeSE takes as input a query $x$ and $N$ responses $R(\cdot|x)=\left\{r^{1}_{T=t}, \ldots, r^{N}_{T=t}\right\}$ generated from $\mathcal{M}$ at temperature $T=t$, and outputs a relative uncertainty score $U(x)$ that quantifies the semantic uncertainty inherent in $\mathcal{M}$'s responses to $x$. LLM-based systems can use $U(x)$ as a quantitative indicator to assess the trustworthiness of $\mathcal{M}$'s response to $x$ and decide whether to abstain in high-uncertainty cases. 

It is important to note that $U(x)$ is not an exact probability of model correctness, the latter pertaining to the orthogonal topic of model calibration \cite{liu2025uncertainty}.
Following prior work, we focus on estimating total uncertainty, which comprises epistemic and aleatoric uncertainty \cite{hou2024decomposing}, as they jointly contribute to hallucinations.

\paragraph{Semantic Uncertainty}
To measure uncertainty at the semantic rather than the token level, \citet{farquhar2024detecting} introduced {Semantic Entropy} (SE). Given an input $x$ and the set of all possible semantic clusters $\Omega$, SE is defined as:
\begin{equation}
    \text{SE}(x) = -\sum_{C \in \Omega} p(C \mid x) \log p(C \mid x).
\end{equation}
Since $\Omega$ is typically intractable, SE is estimated using $M$ clusters $\{C_i\}_{i=1}^M$ extracted from generated samples:
\begin{equation}
    \text{SE}(x) \approx -\sum_{i=1}^{M} p(C_i \mid x) \log p(C_i \mid x),\label{eq:se}
\end{equation}
where $p(C_i \mid x)$ is the normalized semantic probability. When token-level likelihoods are unavailable, $p(C_i \mid x)$ can be approximated by the frequency of samples $S_j \in S$ falling into each cluster: $p(C_i \mid x) \approx \frac{1}{N} \sum_{j=1}^{N} \mathbb{I}(S_j \in C_i)$, which is referred to as {Discrete Semantic Entropy (DSE)}.

\paragraph{Hierarchical Abstraction}
The LLM semantic space models a set of interacting semantic entities and their dynamic relationships within a specific context. Formally, it can be described using a semantic graph $G=(V, E, W)$, where $V$, $E \subseteq V \times V$, and $W:E \to \mathbb{R}_{\geq 0}$ represent the set of sampled responses, directed edges, and interaction strength of edges in the graph, respectively. 
Inspired by the concept of the partitioning tree \cite{li2016structural}, we construct a novel tree structure---{hierarchical abstraction} (Definition \ref{definition:Hierarchical_Abstraction}) to represent the intrinsic hierarchical structure of semantic graph $G$. 

\begin{definition}[Hierarchical Abstraction]\label{definition:Hierarchical_Abstraction}
{The hierarchical abstraction}  of a semantic graph $G= (V, E, W) $ is formalized by an encoding tree $\mathcal{T}$ that satisfies the following conditions:
(1) The root node $\lambda$ whose height is set as 0 contains all nodes in $G$, $\mathcal{T}_\lambda = V$.
Each node $\alpha \in \mathcal{T}$ represents a partition of responses $\mathcal{T}_\alpha \subseteq V$.
For any leaf node $\gamma$, $\mathcal{T}_\gamma$ contains a single response from $V$.
(2) Each non-leaf node $\alpha$ has a nonempty set of immediate successors denoted as $\beta_0, \beta_1, \ldots, \beta_l$. The collection of subsets $\{\mathcal{T}_{\beta_0}, \mathcal{T}_{\beta_1}, \ldots, \mathcal{T}_{\beta_l}\}$ forms a partition of $\mathcal{T}_\alpha$.
\end{definition}

\citep{li2016structural} proposed {structural entropy} for measuring structural information embedded in graph data. Mathematically, the one-dimensional structural entropy of graph $G$ equals to the Shannon entropy of the stationary distribution induced by vertex degrees. It is defined as follows:
\begin{equation}
H^{1}(G)=-\sum_{v \in V} \frac{d_{v}}{{vol}(G)} \log _{2} \frac{d_{v}}{{vol}(G)}, 
\end{equation}
where $d_v$ is the sum of the weights of $v$'s connected edges, and $ {vol}(G)=\sum_{v \in V} d_{v} $ is the volume of $G$. 

The structural entropy of $G$ by a hierarchical abstraction formalized by an encoding tree $\mathcal{T}$ is defined as:
\begin{equation}
H^{\mathcal{T}}(G)= \sum_{\alpha \in \mathcal{T}, \alpha \neq \lambda} H^{\mathcal{T}}(G ; \alpha),
\end{equation}
\begin{equation}
H^{\mathcal{T}}(G ; \alpha)=-\frac{g_{\alpha}}{\text{vol}(G)} \log _{2} \frac{\mathcal{V}_{\alpha}}{\mathcal{V}_{\alpha^{-}}},
\label{eq:entropy_for_node}
\end{equation}
where $g_{\alpha}$ denotes the total weight of edges entering $\mathcal{T}_{\alpha}$ from outside;
$\mathcal{V}_{\alpha}$ denotes the volume of the sub-graph induced by $\mathcal{T}_{\alpha}$, i.e., the sum of degrees of its constituent vertices; and $\alpha^{-}$ denotes the parent node of $\alpha$.

\section{Semantic Structural Entropy}
\label{sec:Method}

\subsection{SeSE for Short-form Generation}\label{sec:SeSE for Short-form Generation}
As shown in Fig.~\ref{fig:SeSE-sentence}, SeSE comprises three phases: response sampling, semantic graph construction and hierarchical abstraction. In Step~3, we construct the optimal semantic hierarchical abstraction based on the structural entropy minimization principle \cite{li2016structural}. The structural entropy of the resulting encoding tree quantifies the inherent uncertainty of the semantic space after optimal compression. Semantic spaces with clear regularities exhibit lower entropy, corresponding to lower uncertainty, while disordered spaces resist compression and yield higher entropy.

\paragraph{Step 1. Response Sampling}
SeSE is a sampling-based approach that requires only a set of black-box responses as input. Given the context $ x $ as input to an LLM $\mathcal{M}$, we sample two types of responses from $\mathcal{M}$: (1) a single greedy-decoded answer $r_{T=0}$ generated at temperature $T=0$, which serves as a proxy for the model's most confident output; and (2) a set of $N$ stochastic samples $R= \left\{r^{1}, \ldots, r^{N}\right\} $ for semantic space modeling.

\paragraph{Step 2. Semantic Graph Construction}
Natural language inference (NLI) models have proven to be effective in analyzing contextual semantic relationships between texts \cite{farquhar2024detecting}.
We employ the NLI model DeBERTa-v3-large-mnli \cite{Deberta} to measure pairwise entailment within the response set $R$. For each ordered pair $(r^i, r^j) \in R \times R$ with $i \neq j$, we construct a premise-hypothesis pair $(x \oplus r^i, x \oplus r^j)$ and feed it into the NLI model. 
We then apply the softmax function $\sigma(\cdot)$ to the predicted logits to obtain a probability distribution:
\begin{equation}
[p_e, p_n, p_c] = \sigma \left( \text{NLI}(x \oplus r^i, x \oplus r^j) \right),
\end{equation}
where $p_e$, $p_n$, and $p_c$ denote the probabilities of entailment, neutrality, and contradiction from the premise to the hypothesis, respectively, conditioned on $x$. Formally, we model the semantic space using a directed weighted graph $G=(V,E,W)$ where the vertex set is $V=R$ and $E$ is the set of directed edges. The weight $W(v_i,v_j)$ is defined as
\begin{equation}
W(v_i,v_j) = p_e+\frac{1}{2} \cdot p_n+ 0 \cdot p_c=p_e+\frac{1}{2} \cdot p_n.
\label{eq:weight function}
\end{equation}
To make $G$ suitable for a random-walk process, we normalize the weights of all outgoing edges by dividing each weight by the vertex's weighted out-degree sum, obtaining the transition matrix $P$:
\begin{equation}
P(v_i,v_j) = \frac{W(v_i,v_j)}{\sum_{(v_i,v_k) \in E} W(v_i,v_k)}.
\label{eq:weight_normalization}
\end{equation}

\paragraph{Step 3. Hierarchical Abstraction}
\begin{algorithm}[tb]
\SetAlgoLined
\caption{Hierarchical Abstraction Construction}
\label{alg:hierarchical_abstracting}
\KwIn{the directed semantic graph $G$, $K>1$ }
\KwOut{the $K$-dimensional optimal encoding tree $\mathcal{T}^*$}
Initialize a one-dimensional encoding tree $\mathcal{T}$ for $G$\\
\While{tree height $h<K$}{
    \ForEach{sibling nodes $\alpha,\beta \in \mathcal{T}$}{
    $\alpha^*,\beta^* \gets \arg \max \Delta_{\alpha, \beta}^{op_{mer}} \text{via Eq.~\ref{eq:entropy variation}}$
    }
    \If{$\Delta_{\alpha^*, \beta^*}^{op_{mer}}>0$}{
    $\mathcal{T} \gets op_{mer}(\mathcal{T},\alpha^*,\beta^*)$,
    continue
    } 
    \ForEach{sibling nodes $\alpha,\beta \in \mathcal{T}$}{
    $\alpha^*,\beta^* \gets \arg \max \Delta_{\alpha, \beta}^{op_{com}} \text{via Eq.~\ref{eq:entropy variation}}$
    }
    \If{$\Delta_{\alpha^*, \beta^*}^{op_{com}}>0$}{
    $\mathcal{T} \gets op_{com}(\mathcal{T},\alpha^*,\beta^*)$, continue
    }
    break
}
\Return $\mathcal{T}^* \gets \mathcal{T}$
\end{algorithm}
Since $P$ is an irreducible row-stochastic matrix with non-negative entries, there exists a unique stationary distribution $\pi$ satisfying $\pi P = \pi$ \cite{norris1998markov}.
We then define and optimize structural entropy for directed graphs based on its stationary distribution, enabling accurate representation of key transition dynamics in semantic spaces. The one-dimensional directed structural entropy ($K=1$) is define as follows:
\begin{equation}
H^1(G) = -\sum_{v \in V} \pi(v) \cdot \log_{2} \pi(v),  
\label{eq:one_directed_se}
\end{equation}
where $\pi(v)$ denotes the stationary probability of vertex $v$. We then refine the definitions of $g_{\alpha}$ and $\mathcal{V}_{\alpha}$ for each non-root node $\alpha$ in the encoding tree $\mathcal{T}$. Following Eq.~\ref{eq:entropy_for_node}, the entropy assigned to $\alpha$ is defined as:
\begin{align}
&\mathcal{V}_{\alpha} = \sum_{v_{i} \in V} \sum_{v_{j} \in V_{\alpha}} \pi(v_{i}) P(v_{i}, v_{j}), \label{eq:refine_V_alpha}\\ 
& g_{\alpha} = \sum_{v_{i} \in V \setminus V_{\alpha}} \sum_{v_{j} \in V_{\alpha}} \pi(v_{i}) P(v_{i}, v_{j}),  \label{eq:refine_g_alpha}\\ 
& H^{\mathcal{T}}(G ; \alpha) = -\frac{g_{\alpha}}{\text{vol}(G)} \log_{2} \frac{\mathcal{V}_{\alpha}}{\mathcal{V}_{\alpha^{-}}} \label{eq:refine_entropy},
\end{align}
where $\text{vol}\left(G\right)=\sum_{v \in V}\left(d_{v}^{+}+d_{v}^{-}\right)$ is the volume of $G$, with $d_{v}^{+}$ and $d_{v}^{-}$ denoting the weighted out-degree and in-degree of vertex $v$, respectively.

The optimal hierarchical abstraction is determined by identifying an encoding tree $\mathcal{T}^*$ with minimal structural entropy \cite{li2024science}. To find $\mathcal{T}^{\star}$, we design an efficient structural entropy minimization algorithm (Algorithm~\ref{alg:hierarchical_abstracting}) using the \emph{merging} ($op_{mer}$) and \emph{combining} ($op_{com}$) operators introduced by \cite{li2024science}. Let $\mathcal{T}_{\alpha,\beta}$ be the encoding tree obtained after executing the merging or combining operator on sibling nodes $\alpha$ and $\beta$ that share a common parent. We define the resulting entropy variation as 
\begin{equation}
\Delta_{\alpha, \beta}^{op}=H^{\mathcal{T}}(G)-H^{\mathcal{T}_{\alpha, \beta}}(G). 
\label{eq:entropy variation}
\end{equation}
Specifically, we first initialize a one-dimensional encoding tree $\mathcal{T}$
: (1) the root node $\lambda$ represents the entire semantic space, i.e., $\mathcal{T}_{\lambda} = R$; (2) for each $r \in R$, we generate a leaf node $\gamma$ with $\mathcal{T}_{\gamma}=\left\{r\right\}$ and assign it as a child node of $\lambda$, i.e., $\gamma^{-} = \lambda$. 
Subsequently, we iteratively optimize the encoding tree $\mathcal{T}$ to $K$ layers. In each iteration, we traverse all sibling node pairs in $\mathcal{T}$ and greedily apply $op_{mer}$ or $op_{com}$, selecting the operation that maximizes the decrease of structural entropy while ensuring that the tree height remains below $K$. The iterative procedure terminates when no sibling pair satisfies $\Delta_{\alpha, \beta}^{op}> 0$ or the tree height reaches $K$, at which point we output $\mathcal{T}^*$. A detailed illustration of the operators and entropy variations is provided in Appendix~\ref{appendix:Hierarchical Abstraction}. Formally, we define SeSE as \emph{the total entropy of the optimal $K$-dimensional encoding tree $\mathcal{T}^*$:}
\begin{align}
&\mathcal{T}^{\star}=\underset{\forall \mathcal{T}: \operatorname{height}(\mathcal{T}) \leq K}{\arg \min }\left(H^{\mathcal{T}}(G)\right),\label{eq:optimal_tree}\\
Se&SE(G)=\sum_{\alpha \in {\mathcal{T}^{\star}}, \alpha \neq \lambda} H^{\mathcal{T}^{\star}}(G ; \alpha)\label{eq:SeSE}.
\end{align}

\paragraph{SeSE Generalizes Semantic Entropy~\citep{farquhar2024detecting}.}
The following theorem shows that SeSE can recover SE for any semantic clustering.
\begin{theorem}[SeSE Generalizes SE]\label{theorem:SeSE generalizes Semantic Entropy}
For any semantic clustering, there exists a semantic graph such that the one-dimensional structural entropy of this graph is equal to semantic entropy (computed as in Eq.~\ref{eq:se}).
\end{theorem}
\vspace{-3mm}
\emph{Proof Sketch.}
When $K=1$, SeSE reduces to the Shannon entropy of the graph's stationary distribution (Eq.~\ref{eq:one_directed_se}). Given any semantic clustering, we can construct a corresponding semantic quotient graph whose stationary distribution exactly matches the clustering distribution, and for which SeSE with $K=1$ equals the SE. Appendix~\ref{appendix:theorem_proofs} provides the detailed proof.

\subsection{SeSE for Long-form Generation}
LLMs often output long-form paragraphs comprising multiple {claims}~\cite{min2023factscore}, which are the smallest semantically distinct information units. In long-form generation, we therefore assess uncertainty at the finer claim level rather than simply assigning a single uncertainty score to an entire response. Given a context $x$, a set of stochastic sampled responses $R$, and a set of claims $C$ extracted from the greedy response $r_{T=0}$, we construct a claim-response bipartite graph $G_{cr}=((R,C),E)$ where an edge $(r,c)\in E$ indicates that response $r$ semantically entails claim $c$.

By minimizing the $K$-dimensional structural entropy of $G_{cr}$ with Algorithm~\ref{alg:hierarchical_abstracting}, we obtain its optimal encoding tree $\mathcal{T}_{cr}^{*}$, which captures the inherent hierarchical community structure over $R\cup C$. For any claim $c \in C$, the uncertainty of reaching $c$ is determined by the cumulative entropy of all non-root nodes $\alpha$ encountered along the path from the root $\lambda$ to the leaf $\gamma$ with $V_{\gamma} = \{c\}$. Accordingly, we define the SeSE of each claim $c$ as \emph{its uncertainty of engaging in random interactions within $G_{cr}$}:
\begingroup
\small
\begin{equation}
\text{SeSE}\left(G_{cr}; c\right) = -\sum_{\alpha \in \mathcal{P}(\lambda \to \gamma) \setminus \{\lambda\}} \frac{g_{\alpha}}{\operatorname{vol}(G_{cr})} \log_{2} \frac{\mathcal{V}_{\alpha}}{\mathcal{V}_{\alpha^{-}}},
\end{equation}
\endgroup
where $\mathcal{P}(\lambda\to\gamma)$ denotes the path from the root $\lambda$ to the leaf $\gamma$ representing $c$. 
Claims with lower SeSE reside in densely connected core regions, reflecting consistent support across sampled responses and lower uncertainty. Conversely, claims with higher SeSE lie in peripheral or sparsely connected regions, indicating a higher uncertainty. Implementation details are provided in Appendix~\ref{appendix:long-form generation}.

\section{Experiments}\label{sec:Experimental Setup}
\subsection{Experimental Setup}
\paragraph{Datasets and LLMs}\label{sec:dataset+llms}
For short-form experiments, we employ five representative free-form QA datasets spanning diverse domains of natural language generation: BioASQ \cite{krithara2023bioasq} (biomedical sciences), SVAMP \cite{patel2021nlp} (mathematical word problems), TriviaQA \cite{joshi2017triviaqa} (trivia knowledge), NQ-Open \cite{nqopen2019} (open-domain natural questions), and SQuAD\_V2 \cite{rajpurkar-etal-2018-know} (reading comprehension). The evaluation is conducted across the Llama-3.1-Instruct series (8B and 70B parameters) \cite{grattafiori2024llama3herdmodels} and the Qwen-3-Instruct series (4B and 30B parameters) \cite{qwen3technicalreport}. Regarding long-form generation, we perform evaluations on two challenging datasets featuring real-world entities from Wikipedia: FActScore \cite{min2023factscore} and PopQA \cite{mallen2023not}. For these benchmarks, we utilize DeepSeek-V3.1 \cite{deepseekai2025deepseekv31} and Gemini-3-Flash \cite{deepmind2025gemini}. Further dataset details are provided in Appendix~\ref{appendix:Dataset}.

\paragraph{Baselines}\label{sec:baselines}
We include a range of widely used baselines for comparison.
In short-form experiments, we evaluate seven representative UQ methods: (1) P(True) \cite{kadavath2022language} uses few-shot prompts to guide LLMs to estimate the probability that their most confident answer is true.
(2) Embedding Regression (ER) \cite{farquhar2024detecting} is a strong supervised baseline that trains a classifier on the final hidden states to predict correctness.
(3) SelfCheckGPT (SC) \cite{manakul2023selfcheckgpt} is a representative self-consistency method.
(4) Length-normalized Predictive Entropy (LN-PE) \cite{malinin2021uncertaintyestimationautoregressivestructured} computes length-normalized token-level log-probabilities.
(5) Semantic Entropy (SE) \cite{farquhar2024detecting} estimates the Shannon entropy over semantic clusters.
(6) Kernel Language Entropy (KLE) \cite{nikitin2024kernel} is a generalization of SE using semantic kernels. 
(7) Semantic Graph Density (SGD) \cite{LiSYJCCR25} quantifies semantic consistency via semantic graph density.
In long-form experiments, in addition to adapting Discrete Semantic Entropy (DSE), SC, and P(True) for long-form generation, we further incorporate two Verbalized Uncertainty methods \cite{mohri2024language}, including Post-hoc Verbalized Uncertainty (PH-VU) and In-line Verbalized Uncertainty (IL-VU). For further details, refer to Appendix \ref{appendix:Baselines}.

\paragraph{Evaluation Metrics}\label{sec:metrics}
Following previous work \cite{farquhar2024detecting}, we assess two primary metrics: Area Under the Receiver Operating Characteristic curve (AUROC) and Area Under the Rejection Accuracy Curve (AURAC) \cite{nadeem2009accuracy}. AUROC evaluates how well the uncertainty scores distinguish between correct and incorrect answers and ranges from 0 to 1, where 1 denotes a perfect classifier and 0.5 indicates random classification. 
AURAC quantifies the potential accuracy improvement users may experience when employing different UQ metrics to exclude high-uncertainty queries. The X\% rejection accuracy represents model performance on the subset of questions retained after filtering out the top X\% high-uncertainty queries, and AURAC provides a comprehensive assessment of accuracy across multiple rejection thresholds.

\paragraph{Implementation Details}
For KLE and SGD, we employ the best variants $\text{KLE}_\text{HEAT}$ and $\text{SGD}_\text{s+P}$, and use DeBERTa-v3-large-mnli as the NLI model, which is identical to SeSE. To ensure fairness, for SC, SE, DSE and SeSE, in long-form experiments, we use identical GPT-5-mini for entailment prediction. 
In all experiments, stochastic responses $ R $ are generated at a temperature of $T=1$ with a size of $N=10$ using nucleus sampling ($P=0.95$) \cite{Holtzman2020The} and top-$K$ sampling ($K=20$) \cite{fan2018hierarchical}. The greedy-decoded response $ r_{T=0} $, obtained at $T=0,P=1,K=20$, serves as the most confident answer and is used for accuracy evaluation. We use GPT-5-mini to automatically evaluate the correctness of $ r_{T=0} $ by comparing it with the reference answer.  
We validate this automated evaluation against human judgment, and relevant results are shown in \textcolor{black}{Appendix}~\ref{sec:automated ground-truth evaluations}.

\subsection{Main Results}
\label{sec:Experimental Results}

\begin{table*}[t]
\caption{Detailed experimental results of \textbf{20} model-dataset pairs in short-form generation. All results are the average of five runs and rounded to two decimal places. In each scenario, the best result is highlighted in \textbf{bolded}. The {Avg./$\Delta^{\%}_{\uparrow}$} presents the LLM-wise percentage improvement of corresponding method compared to the baseline SE.
}
\label{table:sentence_results}
\centering
\begin{threeparttable}
\renewcommand{\arraystretch}{1}
\resizebox{\textwidth}{!}
{
\begin{tabular}{l|l|cc|cc|cc|cc|cc|cc} 
\toprule 
\multicolumn{2}{c}{\raisebox{-1\height}{\textbf{Model/Method}}} &
\multicolumn{2}{c}{\textbf{BioASQ}} & 
\multicolumn{2}{c}{\textbf{NQ-Open}} & 
\multicolumn{2}{c}{\textbf{SQuAD}} & 
\multicolumn{2}{c}{\textbf{SVAMP}} & 
\multicolumn{2}{c}{\textbf{TriviaQA}} & 
\multicolumn{2}{c}{\textbf{Avg./$\Delta^{\%}_{\uparrow}$}} \\ 
\multicolumn{1}{c}{}&\multicolumn{1}{c}{} & \multicolumn{1}{c}{\textbf{AUROC}} & \multicolumn{1}{c}{\textbf{AURAC}} & \multicolumn{1}{c}{\textbf{AUROC}} & \multicolumn{1}{c}{\textbf{AURAC}} & \multicolumn{1}{c}{\textbf{AUROC}} & \multicolumn{1}{c}{\textbf{AURAC}} & \multicolumn{1}{c}{\textbf{AUROC}} & \multicolumn{1}{c}{\textbf{AURAC}} & \multicolumn{1}{c}{\textbf{AUROC}} & \multicolumn{1}{c}{\textbf{AURAC}} & \multicolumn{1}{c}{\textbf{AUROC}} & \multicolumn{1}{c}{\textbf{AURAC}} \\ 
\midrule  
\multirow{8}{*}{\rotatebox{90}{\textbf{Llama-3.1-8B}}} 
&P(True) & 0.68\textcolor{gray}{$\pm$0.02} & 0.56\textcolor{gray}{$\pm$0.03} & 0.68\textcolor{gray}{$\pm$0.02} & 0.43\textcolor{gray}{$\pm$0.03} & 0.63\textcolor{gray}{$\pm$0.03} & 0.27\textcolor{gray}{$\pm$0.03} & 0.68\textcolor{gray}{$\pm$0.06} & 0.60\textcolor{gray}{$\pm$0.03} & {0.76}\textcolor{gray}{$\pm$0.05} & {0.72}\textcolor{gray}{$\pm$0.02} & 0.69 & 0.52\\ 
&ER & {0.67}\textcolor{gray}{$\pm$0.03} & 0.57\textcolor{gray}{$\pm$0.02} & 0.64\textcolor{gray}{$\pm$0.03} & 0.41\textcolor{gray}{$\pm$0.03} & 0.59\textcolor{gray}{$\pm$0.02} & 0.24\textcolor{gray}{$\pm$0.04} & 0.72\textcolor{gray}{$\pm$0.02} & 0.62\textcolor{gray}{$\pm$0.01} & 0.71\textcolor{gray}{$\pm$0.01} & 0.66\textcolor{gray}{$\pm$0.01} &0.66 &0.50 \\ 
&SC & 0.66\textcolor{gray}{$\pm$0.03} & 0.55\textcolor{gray}{$\pm$0.02} & 0.69\textcolor{gray}{$\pm$0.02} & 0.44\textcolor{gray}{$\pm$0.02} & 0.63\textcolor{gray}{$\pm$0.03} & 0.26\textcolor{gray}{$\pm$0.03} & \textbf{0.78}\textcolor{gray}{$\pm$0.03} & \textbf{0.65}\textcolor{gray}{$\pm$0.02} & 0.72\textcolor{gray}{$\pm$0.02} & 0.68\textcolor{gray}{$\pm$0.01} & 0.70 & 0.52 \\ 
&LN-PE & 0.64\textcolor{gray}{$\pm$0.02} & 0.54\textcolor{gray}{$\pm$0.02} & 0.66\textcolor{gray}{$\pm$0.03} & 0.41\textcolor{gray}{$\pm$0.03} & 0.66\textcolor{gray}{$\pm$0.02} & 0.28\textcolor{gray}{$\pm$0.05} & 0.55\textcolor{gray}{$\pm$0.02} & 0.49\textcolor{gray}{$\pm$0.01} & 0.66\textcolor{gray}{$\pm$0.03} & 0.66\textcolor{gray}{$\pm$0.02} & 0.63 &0.48 \\ 
&SE & 0.64\textcolor{gray}{$\pm$0.01} & 0.53\textcolor{gray}{$\pm$0.01} & 0.68\textcolor{gray}{$\pm$0.03} & 0.43\textcolor{gray}{$\pm$0.02} & 0.64\textcolor{gray}{$\pm$0.02} & 0.27\textcolor{gray}{$\pm$0.04} & 0.55\textcolor{gray}{$\pm$0.02} & 0.51\textcolor{gray}{$\pm$0.01} & 0.66\textcolor{gray}{$\pm$0.03} & 0.64\textcolor{gray}{$\pm$0.01} & 0.63$ \;\;\Delta_\text{base}$ & 0.48$ \;\;\Delta_\text{base}$ \\ 
&SGD & 0.66\textcolor{gray}{$\pm$0.02} & 0.56\textcolor{gray}{$\pm$0.03} & 0.70\textcolor{gray}{$\pm$0.01} & 0.45\textcolor{gray}{$\pm$0.02} & 0.64\textcolor{gray}{$\pm$0.02} & 0.28\textcolor{gray}{$\pm$0.04} & 0.57\textcolor{gray}{$\pm$0.03} & 0.50\textcolor{gray}{$\pm$0.02} & 0.68\textcolor{gray}{$\pm$0.02} & 0.66\textcolor{gray}{$\pm$0.01} & 0.65$ \textcolor{orange}{\textbf{+2.4\%}}$ & 0.49$ \textcolor{orange}{\textbf{+3.1\%}}$ \\
&KLE & 0.68\textcolor{gray}{$\pm$0.02} & {0.57}\textcolor{gray}{$\pm$0.02} & 0.71\textcolor{gray}{$\pm$0.02} & 0.46\textcolor{gray}{$\pm$0.02} & 0.67\textcolor{gray}{$\pm$0.02} & 0.30\textcolor{gray}{$\pm$0.04} & 0.66\textcolor{gray}{$\pm$0.02} & 0.58\textcolor{gray}{$\pm$0.02} & 0.76\textcolor{gray}{$\pm$0.03} & 0.71\textcolor{gray}{$\pm$0.01} & 0.69$ \textcolor{orange}{\textbf{+9.4\%}}$ & 0.52$ \textcolor{orange}{\textbf{+10.2\%}}$ \\ 
&\textbf{SeSE(Ours)} & \textbf{0.70}\textcolor{gray}{$\pm$0.02} & \textbf{0.58}\textcolor{gray}{$\pm$0.02} & \textbf{0.77}\textcolor{gray}{$\pm$0.03} & \textbf{0.46}\textcolor{gray}{$\pm$0.03} & \textbf{0.69}\textcolor{gray}{$\pm$0.02} & \textbf{0.35}\textcolor{gray}{$\pm$0.04} & 0.72\textcolor{gray}{$\pm$0.03} & 0.60\textcolor{gray}{$\pm$0.02} & \textbf{0.78}\textcolor{gray}{$\pm$0.01} & \textbf{0.73}\textcolor{gray}{$\pm$0.01} & $\textbf{0.73} \textcolor{orange}{\textbf{+15.3\%}}$ & $\textbf{0.54} \textcolor{orange}{\textbf{+14.4\%}}$ \\ 
\midrule 
\multirow{8}{*}{\rotatebox{90}{\textbf{Llama-3.1-70B}}} 
&P(True) & 0.74\textcolor{gray}{$\pm$0.02} & 0.72\textcolor{gray}{$\pm$0.02} & 0.69\textcolor{gray}{$\pm$0.01} & 0.62\textcolor{gray}{$\pm$0.01} & 0.72\textcolor{gray}{$\pm$0.02} & 0.50\textcolor{gray}{$\pm$0.03} & 0.82\textcolor{gray}{$\pm$0.02} & 0.88\textcolor{gray}{$\pm$0.01} & 0.77\textcolor{gray}{$\pm$0.05} & 0.90\textcolor{gray}{$\pm$0.01} & 0.75 & 0.72 \\ 
&ER & 0.73\textcolor{gray}{$\pm$0.03} & 0.74\textcolor{gray}{$\pm$0.03} & 0.64\textcolor{gray}{$\pm$0.02} & 0.59\textcolor{gray}{$\pm$0.02} & 0.63\textcolor{gray}{$\pm$0.03} & 0.51\textcolor{gray}{$\pm$0.04} & 0.73\textcolor{gray}{$\pm$0.04} & 0.83\textcolor{gray}{$\pm$0.02} & 0.66\textcolor{gray}{$\pm$0.03} & 0.88\textcolor{gray}{$\pm$0.01} &0.68 &0.71 \\ 
&SC & 0.75\textcolor{gray}{$\pm$0.02} & 0.72\textcolor{gray}{$\pm$0.02} & 0.69\textcolor{gray}{$\pm$0.03} & 0.62\textcolor{gray}{$\pm$0.03} & 0.81\textcolor{gray}{$\pm$0.03} & 0.56\textcolor{gray}{$\pm$0.04} & 0.86\textcolor{gray}{$\pm$0.02} & 0.90\textcolor{gray}{$\pm$0.01} & {0.81}\textcolor{gray}{$\pm$0.04} & 0.91\textcolor{gray}{$\pm$0.01} & 0.78 & 0.74 \\ 
&LN-PE & 0.74\textcolor{gray}{$\pm$0.02} & 0.72\textcolor{gray}{$\pm$0.03} & 0.70\textcolor{gray}{$\pm$0.03} & 0.63\textcolor{gray}{$\pm$0.02} & 0.70\textcolor{gray}{$\pm$0.02} & 0.48\textcolor{gray}{$\pm$0.04} & 0.73\textcolor{gray}{$\pm$0.03} & 0.86\textcolor{gray}{$\pm$0.01} & 0.74\textcolor{gray}{$\pm$0.02} & 0.90\textcolor{gray}{$\pm$0.01} & 0.72 &0.72 \\ 
&SE & 0.79\textcolor{gray}{$\pm$0.02} & 0.74\textcolor{gray}{$\pm$0.02} & 0.72\textcolor{gray}{$\pm$0.03} & 0.64\textcolor{gray}{$\pm$0.02} & 0.78\textcolor{gray}{$\pm$0.04} & 0.53\textcolor{gray}{$\pm$0.04} & 0.83\textcolor{gray}{$\pm$0.03} & 0.88\textcolor{gray}{$\pm$0.01} & 0.76\textcolor{gray}{$\pm$0.05} & 0.90\textcolor{gray}{$\pm$0.01} & 0.78$ \;\;\Delta_\text{base}$ & 0.74$ \;\;\Delta_\text{base}$ \\ 
&SGD & 0.81\textcolor{gray}{$\pm$0.02} & 0.77\textcolor{gray}{$\pm$0.03} & 0.73\textcolor{gray}{$\pm$0.04} & 0.65\textcolor{gray}{$\pm$0.03} & 0.78\textcolor{gray}{$\pm$0.03} & 0.53\textcolor{gray}{$\pm$0.04} & 0.85\textcolor{gray}{$\pm$0.00} & 0.89\textcolor{gray}{$\pm$0.01} & 0.78\textcolor{gray}{$\pm$0.05} & 0.91\textcolor{gray}{$\pm$0.01} & 0.79$ \textcolor{orange}{\textbf{+1.9\%}}$ & 0.75$ \textcolor{orange}{\textbf{+1.6\%}}$ \\
&KLE & {0.83}\textcolor{gray}{$\pm$0.02} & 0.77\textcolor{gray}{$\pm$0.02} & 0.75\textcolor{gray}{$\pm$0.03} & 0.66\textcolor{gray}{$\pm$0.02} & 0.79\textcolor{gray}{$\pm$0.02} & 0.54\textcolor{gray}{$\pm$0.04} & 0.87\textcolor{gray}{$\pm$0.02} & 0.89\textcolor{gray}{$\pm$0.01} & 0.81\textcolor{gray}{$\pm$0.03} & {0.91}\textcolor{gray}{$\pm$0.01} & 0.81$ \textcolor{orange}{\textbf{+4.2\%}}$ & 0.75$ \textcolor{orange}{\textbf{+2.2\%}}$\\ 
&\textbf{SeSE(Ours)} & \textbf{0.84}\textcolor{gray}{$\pm$0.01} & \textbf{0.77}\textcolor{gray}{$\pm$0.02} & \textbf{0.80}\textcolor{gray}{$\pm$0.02} & \textbf{0.70}\textcolor{gray}{$\pm$0.01} & \textbf{0.83}\textcolor{gray}{$\pm$0.03} & \textbf{0.57}\textcolor{gray}{$\pm$0.04} & \textbf{0.88}\textcolor{gray}{$\pm$0.01} & \textbf{0.94}\textcolor{gray}{$\pm$0.01} & \textbf{0.82}\textcolor{gray}{$\pm$0.04} & \textbf{0.91}\textcolor{gray}{$\pm$0.01} & $\textbf{0.83} \textcolor{orange}{\textbf{+7.3\%}}$ & $\textbf{0.78} \textcolor{orange}{\textbf{+5.4\%}}$\\ 
\midrule 
\multirow{8}{*}{\rotatebox{90}{\textbf{Qwen-3-4B}}} 
&P(True) & 0.75\textcolor{gray}{$\pm$0.02} & {0.67}\textcolor{gray}{$\pm$0.03} & 0.80\textcolor{gray}{$\pm$0.08} & {0.44}\textcolor{gray}{$\pm$0.02} & 0.75\textcolor{gray}{$\pm$0.04} & 0.39\textcolor{gray}{$\pm$0.06} & 0.85\textcolor{gray}{$\pm$0.01} & 0.84\textcolor{gray}{$\pm$0.01} & 0.83\textcolor{gray}{$\pm$0.05} & 0.74\textcolor{gray}{$\pm$0.02} & 0.79 & 0.61 \\ 
&ER & 0.72\textcolor{gray}{$\pm$0.02} & {0.68}\textcolor{gray}{$\pm$0.01} & 0.72\textcolor{gray}{$\pm$0.05} & 0.35\textcolor{gray}{$\pm$0.03} & 0.64\textcolor{gray}{$\pm$0.05} & 0.33\textcolor{gray}{$\pm$0.04} & 0.80\textcolor{gray}{$\pm$0.02} & 0.80\textcolor{gray}{$\pm$0.00} & 0.76\textcolor{gray}{$\pm$0.03} & {0.78}\textcolor{gray}{$\pm$0.02} &0.73 &0.59 \\ 
&SC & 0.74\textcolor{gray}{$\pm$0.01} & 0.66\textcolor{gray}{$\pm$0.01} & 0.76\textcolor{gray}{$\pm$0.02} & 0.40\textcolor{gray}{$\pm$0.02} & 0.75\textcolor{gray}{$\pm$0.03} & 0.40\textcolor{gray}{$\pm$0.04} & {0.87}\textcolor{gray}{$\pm$0.02} & 0.85\textcolor{gray}{$\pm$0.01} & 0.78\textcolor{gray}{$\pm$0.03} & 0.72\textcolor{gray}{$\pm$0.02} & 0.78 & 0.60 \\ 
&LN-PE & 0.74\textcolor{gray}{$\pm$0.04} & 0.66\textcolor{gray}{$\pm$0.03} & 0.77\textcolor{gray}{$\pm$0.06} & 0.42\textcolor{gray}{$\pm$0.03} & 0.77\textcolor{gray}{$\pm$0.06} & {0.42}\textcolor{gray}{$\pm$0.07} & 0.70\textcolor{gray}{$\pm$0.06} & 0.76\textcolor{gray}{$\pm$0.02} & 0.78\textcolor{gray}{$\pm$0.05} & {0.78}\textcolor{gray}{$\pm$0.02} & 0.76 &0.60 \\ 
&SE & 0.73\textcolor{gray}{$\pm$0.03} & 0.65\textcolor{gray}{$\pm$0.01} & 0.79\textcolor{gray}{$\pm$0.04} & 0.43\textcolor{gray}{$\pm$0.02} & 0.77\textcolor{gray}{$\pm$0.03} & 0.41\textcolor{gray}{$\pm$0.04} & 0.70\textcolor{gray}{$\pm$0.01} & 0.77\textcolor{gray}{$\pm$0.01} & 0.82\textcolor{gray}{$\pm$0.04} & 0.73\textcolor{gray}{$\pm$0.02} & 0.76$ \;\;\Delta_\text{base}$ & 0.60$ \;\;\Delta_\text{base}$ \\ 
&SGD & 0.78\textcolor{gray}{$\pm$0.03} & 0.66\textcolor{gray}{$\pm$0.03} & \textbf{0.83}\textcolor{gray}{$\pm$0.04} & \textbf{0.46}\textcolor{gray}{$\pm$0.02} & 0.77\textcolor{gray}{$\pm$0.01} & 0.42\textcolor{gray}{$\pm$0.03} & 0.72\textcolor{gray}{$\pm$0.02} & 0.80\textcolor{gray}{$\pm$0.01} & 0.85\textcolor{gray}{$\pm$0.03} & 0.76\textcolor{gray}{$\pm$0.02} & 0.79$ \textcolor{orange}{\textbf{+3.4\%}}$ & 0.62$ \textcolor{orange}{\textbf{+3.1\%}}$ \\
&KLE & 0.78\textcolor{gray}{$\pm$0.02} & 0.67\textcolor{gray}{$\pm$0.02} & {0.80}\textcolor{gray}{$\pm$0.03} & 0.43\textcolor{gray}{$\pm$0.02} & {0.79}\textcolor{gray}{$\pm$0.03} & 0.40\textcolor{gray}{$\pm$0.04} & 0.79\textcolor{gray}{$\pm$0.02} & 0.81\textcolor{gray}{$\pm$0.01} & 0.85\textcolor{gray}{$\pm$0.04} & 0.75\textcolor{gray}{$\pm$0.02} & 0.80$ \textcolor{orange}{\textbf{+5.4\%}}$ & 0.61$ \textcolor{orange}{\textbf{+2.7\%}}$\\ 
&\textbf{SeSE(Ours)} & \textbf{0.79}\textcolor{gray}{$\pm$0.02} & \textbf{0.68}\textcolor{gray}{$\pm$0.01} & {0.81}\textcolor{gray}{$\pm$0.02} & {0.44}\textcolor{gray}{$\pm$0.02} & \textbf{0.80}\textcolor{gray}{$\pm$0.02} & \textbf{0.42}\textcolor{gray}{$\pm$0.04} & \textbf{0.87}\textcolor{gray}{$\pm$0.02} & \textbf{0.86}\textcolor{gray}{$\pm$0.01} & \textbf{0.87}\textcolor{gray}{$\pm$0.04} & \textbf{0.78}\textcolor{gray}{$\pm$0.01} & $\textbf{0.83} \textcolor{orange}{\textbf{+8.5\%}}$ & $\textbf{0.63} \textcolor{orange}{\textbf{+6.2\%}}$\\ 
\midrule 
\multirow{8}{*}{\rotatebox{90}{\textbf{Qwen-3-30B-A3B}}} 
&P(True) & 0.70\textcolor{gray}{$\pm$0.02} & 0.70\textcolor{gray}{$\pm$0.05} & {0.80}\textcolor{gray}{$\pm$0.03} & 0.57\textcolor{gray}{$\pm$0.06} & 0.68\textcolor{gray}{$\pm$0.03} & 0.43\textcolor{gray}{$\pm$0.09} & \textbf{0.90}\textcolor{gray}{$\pm$0.04} & \textbf{0.89}\textcolor{gray}{$\pm$0.03} & 0.84\textcolor{gray}{$\pm$0.05} & {0.87}\textcolor{gray}{$\pm$0.03} & 0.79 & 0.69 \\ 
&ER & {0.76}\textcolor{gray}{$\pm$0.03} & 0.70\textcolor{gray}{$\pm$0.03} & 0.67\textcolor{gray}{$\pm$0.01} & 0.54\textcolor{gray}{$\pm$0.06} & 0.64\textcolor{gray}{$\pm$0.02} & {0.48}\textcolor{gray}{$\pm$0.08} & 0.85\textcolor{gray}{$\pm$0.02} & 0.84\textcolor{gray}{$\pm$0.02} & 0.73\textcolor{gray}{$\pm$0.05} & 0.83\textcolor{gray}{$\pm$0.03} &0.73 &0.68 \\ 
&SC & 0.74\textcolor{gray}{$\pm$0.01} & 0.71\textcolor{gray}{$\pm$0.05} & 0.77\textcolor{gray}{$\pm$0.02} & 0.54\textcolor{gray}{$\pm$0.05} & 0.68\textcolor{gray}{$\pm$0.02} & 0.42\textcolor{gray}{$\pm$0.10} & {0.89}\textcolor{gray}{$\pm$0.03} & {0.89}\textcolor{gray}{$\pm$0.03} & \textbf{0.86}\textcolor{gray}{$\pm$0.05} & \textbf{0.87}\textcolor{gray}{$\pm$0.03} & 0.79 & 0.69 \\ 
&LN-PE & 0.70\textcolor{gray}{$\pm$0.03} & 0.70\textcolor{gray}{$\pm$0.05} & 0.75\textcolor{gray}{$\pm$0.01} & 0.53\textcolor{gray}{$\pm$0.06} & 0.69\textcolor{gray}{$\pm$0.03} & 0.43\textcolor{gray}{$\pm$0.09} & 0.68\textcolor{gray}{$\pm$0.02} & 0.82\textcolor{gray}{$\pm$0.02} & 0.68\textcolor{gray}{$\pm$0.05} & 0.81\textcolor{gray}{$\pm$0.03} & 0.70 &0.66\\ 
&SE & 0.73\textcolor{gray}{$\pm$0.02} & 0.71\textcolor{gray}{$\pm$0.03} & 0.76\textcolor{gray}{$\pm$0.01} & 0.53\textcolor{gray}{$\pm$0.06} & 0.75\textcolor{gray}{$\pm$0.02} & 0.46\textcolor{gray}{$\pm$0.09} & 0.83\textcolor{gray}{$\pm$0.02} & 0.87\textcolor{gray}{$\pm$0.03} & 0.73\textcolor{gray}{$\pm$0.05} & 0.82\textcolor{gray}{$\pm$0.03} & 0.75$ \;\;\Delta_\text{base}$ & 0.68$ \;\;\Delta_\text{base}$ \\ 
&SGD & 0.73\textcolor{gray}{$\pm$0.03} & 0.73\textcolor{gray}{$\pm$0.05} & 0.78\textcolor{gray}{$\pm$0.04} & 0.54\textcolor{gray}{$\pm$0.08} & 0.75\textcolor{gray}{$\pm$0.03} & \textbf{0.49}\textcolor{gray}{$\pm$0.09} & 0.83\textcolor{gray}{$\pm$0.03} & 0.86\textcolor{gray}{$\pm$0.03} & 0.76\textcolor{gray}{$\pm$0.05} & 0.83\textcolor{gray}{$\pm$0.03} & 0.77$ \textcolor{orange}{\textbf{+2.2\%}}$ & 0.69$ \textcolor{orange}{\textbf{+2.0\%}}$ \\
&KLE & 0.75\textcolor{gray}{$\pm$0.02} & 0.72\textcolor{gray}{$\pm$0.05} & 0.79\textcolor{gray}{$\pm$0.02} & 0.56\textcolor{gray}{$\pm$0.05} & {0.76}\textcolor{gray}{$\pm$0.02} & {0.47}\textcolor{gray}{$\pm$0.10} & 0.84\textcolor{gray}{$\pm$0.02} & 0.88\textcolor{gray}{$\pm$0.02} & 0.78\textcolor{gray}{$\pm$0.05} & 0.84\textcolor{gray}{$\pm$0.03} & 0.79$ \textcolor{orange}{\textbf{+4.3\%}}$ & 0.69$ \textcolor{orange}{\textbf{+2.4\%}}$\\ 
&\textbf{SeSE(Ours)} & \textbf{0.76}\textcolor{gray}{$\pm$0.01} & \textbf{0.75}\textcolor{gray}{$\pm$0.04} & \textbf{0.80}\textcolor{gray}{$\pm$0.01} & \textbf{0.58}\textcolor{gray}{$\pm$0.05} & \textbf{0.76}\textcolor{gray}{$\pm$0.03} & {0.48}\textcolor{gray}{$\pm$0.10} & 0.88\textcolor{gray}{$\pm$0.02} & 0.87\textcolor{gray}{$\pm$0.02} & {0.79}\textcolor{gray}{$\pm$0.04} & {0.87}\textcolor{gray}{$\pm$0.02} & $\textbf{0.80} \textcolor{orange}{\textbf{+6.1\%}}$ & $\textbf{0.71} \textcolor{orange}{\textbf{+5.0\%}}$\\ 
\bottomrule 
\end{tabular}
}
\end{threeparttable}
\end{table*}

\begin{table}[!ht]
  \caption{Detailed experimental results of \textbf{4} model-dataset pairs in long-form generation. 
 $\text{Abs}.(\%) \uparrow$ presents the percentage improvement of \textbf{bolded} method compared to \underline{underlined} method.}
  \label{table:longform_results}
  \centering
    \renewcommand{\arraystretch}{1} 
    \resizebox{\linewidth}{!}{
    \begin{tabular}{l|lcccc}
      \toprule
       \multicolumn{2}{c}{\raisebox{-1\height}{\textbf{Model/Method}}} & \multicolumn{2}{c}{\textbf{FActScore}} & \multicolumn{2}{c}{\textbf{PopQA}}   \\
      \multicolumn{2}{c}{} &
      \textbf{AUROC} & \textbf{AURAC} & \textbf{AUROC} & \textbf{AURAC}\\
      \midrule
      \multirow{7}{*}{\rotatebox{90}{\textbf{DeepSeek-V3.1}}} 
      & IL-VU & 0.5380 & 0.6281 & 0.5198 & 0.6394 \\
      & PH-VU & 0.6366 & 0.7060 & 0.6476 & 0.7307 \\
      & SC & 0.6066 & 0.6864 & 0.6210 & 0.7331 \\
      & P(True) & 0.7216 & 0.7371 & 0.7247 & 0.7709 \\
      & DSE & \underline{0.7842} & \underline{0.7684} & \underline{0.7909} & \underline{0.8094} \\
      \cmidrule{2-6}
      & \textbf{SeSE(Ours)} & \textbf{0.8105} & \textbf{0.7801} & \textbf{0.8468} & \textbf{0.8224}\\
      & $\text{Abs}.(\%)$ &  \textcolor{orange}{$\uparrow \textbf{3.35\%}$} &\textcolor{orange}{$\uparrow \textbf{1.52\%}$} &\textcolor{orange}{$\uparrow \textbf{7.07\%}$} &\textcolor{orange}{$\uparrow \textbf{1.61\%}$}\\
      \midrule
      \multirow{7}{*}{\rotatebox{90}{\textbf{Gemini-3-Flash}}} 
      & IL-VU & 0.5260 & 0.6599 & 0.5823 & 0.6850 \\
      & PH-VU & 0.6713 & 0.7210 & 0.6753 & 0.7300 \\
      & SC & 0.6226 & 0.6874 & 0.6461 & 0.7006 \\
      & P(True) & 0.7658 & 0.7800 & 0.7891 & 0.7791 \\
      & DSE & \underline{0.8315} & \underline{0.8057}& \underline{0.8480} & \underline{0.8055} \\
      \cmidrule{2-6}
      & \textbf{SeSE(Ours)} & \textbf{0.8581} & \textbf{0.8180} & \textbf{0.8588} & \textbf{0.8119}\\
      & $\text{Abs}.(\%) \uparrow$ 
      & \textcolor{orange}{$\uparrow \textbf{3.20\%}$} &\textcolor{orange}{$\uparrow \textbf{1.53\%}$} &\textcolor{orange}{$\uparrow \textbf{1.27}\%$} &\textcolor{orange}{$\uparrow \textbf{0.79}\%$}\\
      \bottomrule
    \end{tabular}
}
\vspace{-3mm}
\end{table}

\paragraph{Short-form Results}
Table~\ref{table:sentence_results} summarizes the short-form experimental results across various datasets and LLMs. The results show that SeSE consistently outperforms strong white-box method SGD and the supervised baseline ER across different model families (Llama-3.1 and Qwen-3) and parameter scales (ranging from 3B to 70B). In particular, with Llama-3.1-8B, SeSE achieves improvements of {15.3\%} in AUROC and {14.4\%} in AURAC compared with the SE baseline.
On average across the five LLMs, SeSE significantly surpasses all entropy-based baselines. Compared to KLE, a recent refinement of SE, SeSE achieves an average improvement of {3.5\%} in AUROC and {3.0\%} in AURAC. The LN-PE performs the worst, as it computes average predictive entropy solely from token-sequence probabilities, thereby conflating semantic and lexical uncertainty. Although KLE attempts to use the von Neumann graph entropy of semantic graph kernels to mitigate the one-cut semantic equivalence limitation of SE, it still performs semantic analysis at a single, non-hierarchical abstraction level. In contrast, SeSE constructs hierarchical semantic abstractions, enabling more precise modeling of uncertainty across multiple semantic levels and providing finer discrimination of nuanced uncertainties. Importantly, SeSE works in a zero-resource manner, which does not require external databases or additional training. The comparison of computational cost are provided in Appendix~\ref{sec:compare_resource}

\vspace{-2mm}
\paragraph{Long-form Results}
We find that even the most advanced LLMs exhibit significant hallucination rates in our custom long-form datasets (including {7,407} claims), with 28\% for DeepSeek-V3.1 and 25\% for Gemini-3-Flash, on average. As shown in Table~\ref{table:longform_results}, SeSE effectively identifies hallucinations in long-form generation with superior AUROC and AURAC scores compared to all baselines, including higher-cost methods such as SC. For DeepSeek-V3.1, SeSE delivers average improvements of {5.21\%} in AUROC and {1.57\%} in AURAC over the second-best DSE. Additionally, verbalized uncertainty performs poorly, suggesting contemporary LLMs remain overconfident even when they should be uncertain about the factuality of their responses. Therefore, reliable uncertainty estimators are crucial for building user trust in LLMs and mitigating deployment risks in high-stakes scenarios. For further results, refer to Appendix~\ref{appendix:Additional Experimental Details}.

\begin{figure}[htbp]
	\centering
\includegraphics[width=\linewidth]{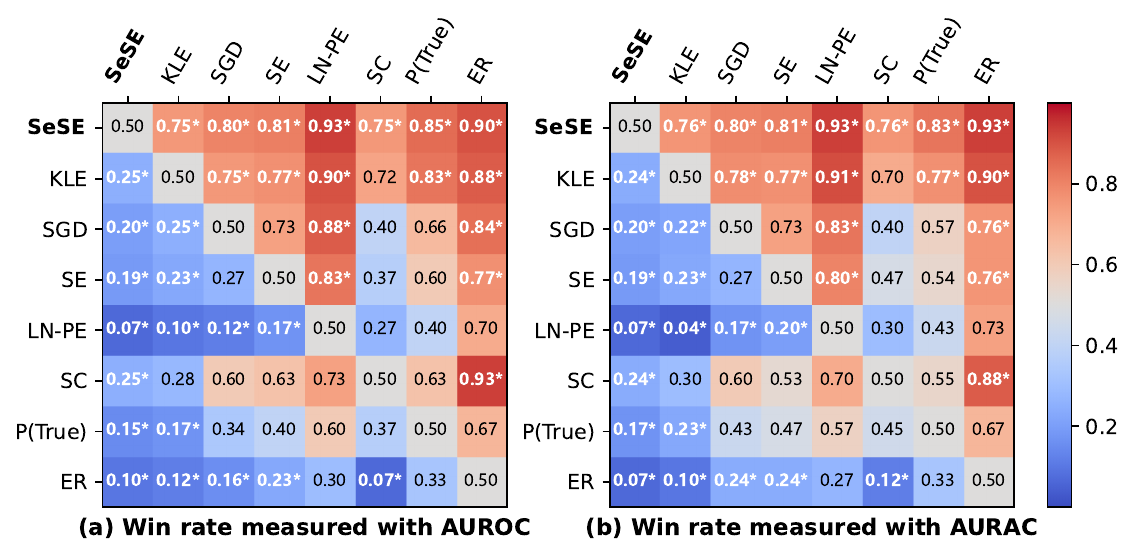}
	\caption{Pairwise win rates across 20 model-dataset scenarios. 
     Each cell indicates the proportion of scenarios where the row method outperforms the column method. 
    Green value with a asterisk ($*$) indicates the binomial statistical significance level $p < 0.05$ according.}
	\label{win_rate_heatmaps}
 \vspace{-3mm}
\end{figure}

\paragraph{Statistical Significance}  
While standard errors are reported in Table~\ref{table:sentence_results}, it should be noted that in the context of LLM UQ, standard errors tend to depend more on the LLM and the dataset than on the UQ method itself \cite{farquhar2024detecting}. Rather than absolute values, the consistency of relative results across scenarios serves as a more reliable indicator of performance variation \cite{farquhar2024detecting}. 
Therefore, we perform a binomial statistical significance test across all baselines. First, we conduct five repeated experiments with distinct random seeds across 20 experimental scenarios (100 runs). For each run, we compute the 95\% confidence interval using 1,000 bootstrap resamples. We then adopt the pairwise win rate as the primary evaluation metric. Within each scenario, the method with more wins across five repeated trials is deemed superior. The heatmaps in Figure~\ref{win_rate_heatmaps} visualize the pairwise win rates, showing that SeSE consistently outperforms baselines at a significance level of $p < 0.05$. This finding indicates that although the performance of SeSE may vary across scenarios, its comparative advantage remains highly consistent, making it a more stable uncertainty estimator in practical applications.

\begin{figure*}[htb]
    \centering
    \begin{subfigure}[t]{0.48\linewidth}
        \centering
        \includegraphics[width=\linewidth]{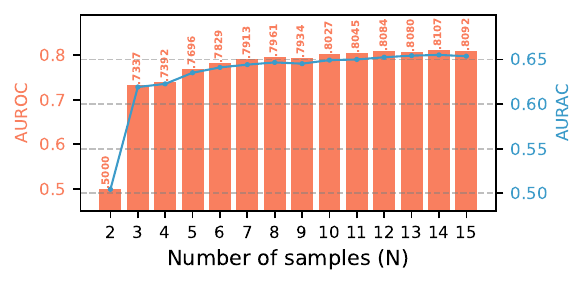}
        \caption{Number of short-form generations used for entropy.}
        \label{fig:Number of short-form generations}
    \end{subfigure}
    \hfill
    \begin{subfigure}[t]{0.48\linewidth}
        \centering
        \includegraphics[width=\linewidth]{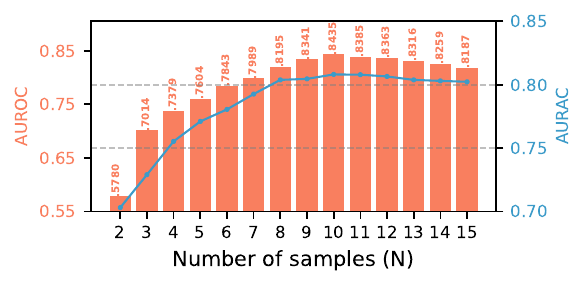}
        \caption{Number of long-form generations used for entropy.}
        \label{fig:Number of long-form generations}
    \end{subfigure}
    
    \caption{The performance of SeSE with different number of sampled responses ($N$). 
    }
    \label{fig:sensitivity to N}
\end{figure*}

\begin{figure*}[htb]
	\centering
	\includegraphics[width=\linewidth]{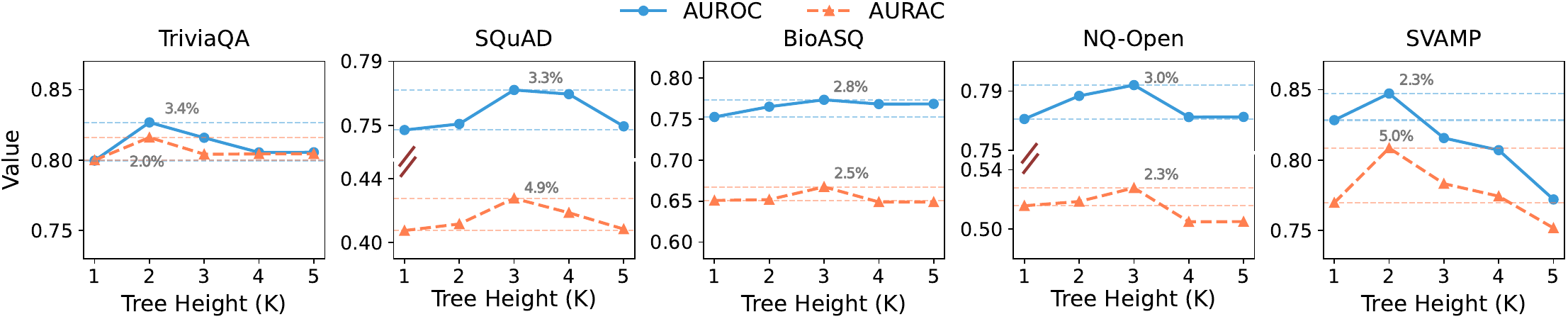}
	\caption{The performance of SeSE when adopting different tree height $K$.}
	\label{fig:K_sensitivity}
 \vspace{-3mm}
\end{figure*}

\subsection{Ablation Experiments}\label{sec:Sensitivity to Hyperparameters}
\vspace{-2mm}
\paragraph{Number of Sampling Numbers}\label{sec:Sensitivity to N}
The number of sampled responses $N$ does not need to be large. Figure~\ref{fig:sensitivity to N} illustrates how the performance of uncertainty quantification varies with $N$ for both short-form and long-form generations. The reported values are aggregated across all datasets and LLMs. For short-form generations, performance gains show diminishing returns at $N \approx 5$. However, increasing $N$ to 10 can still be beneficial. For long-form generation, we find that optimal performance is achieved with $9\text{-}10$ responses. Different from short-form scenarios, more generations don't always improve performance. This occurs because the stochastic decoding strategy of LLMs increases the likelihood of selecting low-probability tokens. Excessively increasing sample numbers amplifies the selection probability of selecting such tokens, potentially introducing irrelevant content and thereby diminishing the relative weight of original greedily-decoding claims. We provide unaggregated results for Figure~\ref{fig:sensitivity to N} in Appendix~\ref{sec:Detailed results of hyperparameter sensitivity}.

\paragraph{Encoding Tree Height}\label{sec:Sensitivity to K}
Figure~\ref{fig:K_sensitivity} shows SeSE's performance sensitivity to the encoding tree height $K$. The accuracy improvement of the best tree height compared with $K=1$ is annotated in the figure. When $K=1$, SeSE degenerates to the non-hierarchical Shannon entropy of the graph's stationary distribution. Compared with $K=1$, SeSE achieves its optimal performance on two and three datasets at $K=2$ and $K=3$, respectively. We also observe that the optimal $K$ correlates with task difficulty. For instance, on the simpler dataset TriviaQA, SeSE peaks at $K=2$, while on the challenging SQuAD, the optimum is found at $K=3$. These findings demonstrate that SeSE can flexibly adapt to diverse downstream tasks with optional encoding tree depth. Additional statistics and detailed analyses for Figure~\ref{fig:K_sensitivity} are provided in Appendix~\ref{sec:Detailed results of hyperparameter sensitivity}.

\section{Related Work}
\label{sec:relat}
\paragraph{Short-form Uncertainty Estimation in LLMs} 
Recently, numerous UQ methods have emerged. A primary direction involves supervised learning, either by fine-tuning base LLMs or adding external layers to predict uncertainty scores \cite{liu2024can,xie2024calibrating,li-etal-2025-know}. Despite their promise, these methods are typically model-specific and cannot be applied to closed-source models, resulting in limited scalability and availability. 
Another line of research explores verbalized uncertainty, prompting LLMs to express uncertainty via natural language \cite{kadavath2022language,tian2023just,mohri2024language,wang2025sconuselectiveconformaluncertainty}. However, since most existing evaluation methods fail to incentivize models to express uncertainty honestly \cite{kalai2025language}, LLMs tend to exhibit overconfidence even when generating incorrect outputs. Consequently, verbalized uncertainty underperforms probabilistic methods \cite{mohri2024language}.

The aforementioned methods primarily focus on lexical uncertainty, neglecting semantic uncertainty, which is a more essential indicator of LLM trustworthiness as it directly reflects response correctness \cite{farquhar2024detecting}. Semantic Entropy \cite{farquhar2024detecting} represents a significant advance that calculates the Shannon entropy of semantic equivalence clusters as an uncertainty metric. However, SE is a binary, one-cut measurement that overlooks finer semantic differences between responses. While recent follow-ups like Kernel Language Entropy (KLE) \cite{nikitin2024kernel} and Semantic Graph Density (SGD) \cite{XiaoBGMMPGL25} model fine-grained semantic relationships in semantic graphs, they fail to capture the semantic topological structure which reflects the informational essence of graphs \cite{li2024survey,SuP0L25}, limiting their ability to distinguish between fine-grained uncertainties.
To address this, SeSE quantifies the uncertainty inherent in the semantic graph after optimal compression based on the structural entropy minimization principle, providing more precise and interpretable uncertainty estimates. 

\vspace{-2mm}
\paragraph{Granular uncertainty estimation} 
Quantifying uncertainty in long-form generation has attracted increasing attention \citep{manakul2023selfcheckgpt,mohri2024language,zhang-etal-2024-luq,jiang2024graph}. Most relevant to our work is Graph Uncertainty \citep{jiang2024graph}, which employs graph centrality metrics like degree and closeness as heuristic proxies for uncertainty. While SeSE also operates on a bipartite semantic graph in long-form settings, the key distinction is that SeSE exploits latent semantic structural dependencies from a random walk perspective, providing interpretable uncertainty estimates for black-box LLMs.

\vspace{-2mm}
\paragraph{Structural Information Theory}
Structural information theory \cite{li2016structural} provides a framework to quantify dynamic uncertainty in complex networks. Minimizing structural entropy, by searching for a nested partitioning tree, provides a theoretically grounded method to identify the intrinsic community structure of a network \cite{li2024science}. This theory has been successfully applied across diverse domains, including graph learning \cite{20244917489114}, social networks \cite{yang2024sebot}, and reinforcement learning \cite{Zeng0L24,11314763}.

\section{Conclusion}
\label{sec:conclusion}
In this paper, we propose SeSE, a principled black-box UQ framework that works for both open- and closed-source LLMs. By constructing the optimal hierarchical abstraction, SeSE quantifies uncertainty inherent in the semantic space after optimal compression, serving as an expressive generalization of semantic entropy. Furthermore, it offers interpretable and fine-grained uncertainty estimates for long-form LLM generation. Extensive experiments show that SeSE outperforms baseline methods. These findings highlight the potential of SeSE for assessing LLM trustworthiness. Future work may explore extending SeSE to multi-agent systems and multi-modal LLMs.

\subsubsection*{Acknowledgements}
This work was supported by the Beijing Natural Science Foundation under Grant L253021, in part by NSFC under Grant 62322202 and Grant U25B2029, in part by the Pioneer and Leading Goose R\&D Program of Zhejiang through grant 2025C02044, in part by the Local Science and Technology Development Fund of Hebei Province Guided by the Central Government of China under Grant 254Z9902G, and in part by the Science Research Project of Hebei Higher Education Institutions under grant CYZD2026005, in part by Shijiazhuang Science and Technology Plan Project under Grant 2511301807A, and in part by CCF-DiDi GAIA collaborative Research Funds for Young Scholars through grant 202527.

\bibliography{cite}

\onecolumn
\emptythanks

\title{SeSE: Black-Box Uncertainty Quantification for \\Large Language Models Based on Structural Information Theory \\(Supplementary Material)}
\maketitle

\vspace{1em}
\appendix
\startcontents[appendices]

\printcontents[appendices]{l}{1}{\setcounter{tocdepth}{2}}

\section{Worked Example of SeSE}
\label{appendix:worked_example}

\begin{figure}[htbp]
    \centering
\includegraphics[width=0.8\linewidth]{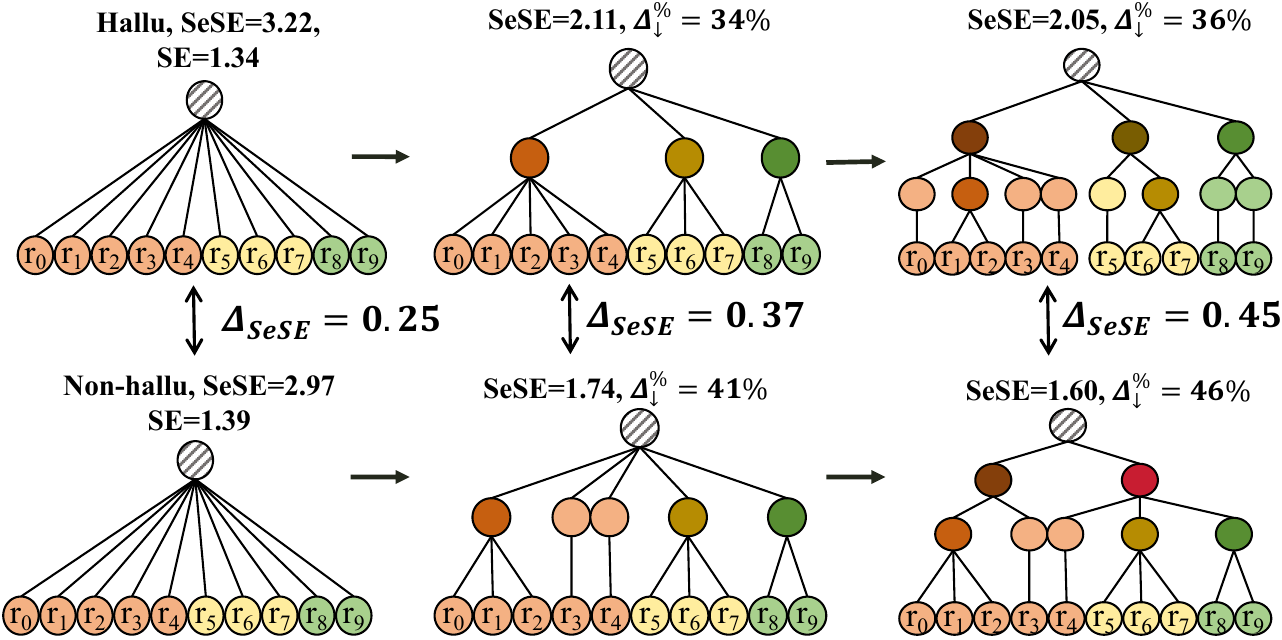}
    \caption{Visualization of the constructed encoding trees for Llama-3.1-8B (Non-hallucinatory) and Llama-3.2-3B (Hallucinatory). SeSE captures the hierarchical semantic structures, assigning a lower structural entropy to the consistent answers of Llama-3.1-8B and a higher entropy to the contradictory answers of Llama-3.2-3B.}
    \label{fig:sese_worked_example}
\end{figure}

To demonstrate what SeSE captures in practice, we provide a real example from the SQuAD dataset. This example illustrates the constructed encoding trees and shows exactly how SeSE captures hierarchical semantic structures that existing methods miss.

\paragraph{Setup}
Consider a question: ``Where was Albert Einstein born?'' and the reference answer is ``Ulm, Germany''. We sample 10 responses from two models: Llama-3.1-8B (non-hallucinatory but varying in granularity) and Llama-3.2-3B (hallucinatory). The sampled responses and their corresponding semantic cluster probabilities are summarized in Table~\ref{tab:worked_example_samples}.

\begin{table}[htbp]
\centering
\caption{Sampled responses from two LLMs for the question ``Where was Albert Einstein born?''.}
\label{tab:worked_example_samples}
\resizebox{0.8\textwidth}{!}{
\begin{tabular}{cll}
\toprule
\textbf{Samples} & \textbf{Model A: Llama-3.1-8B (Non-hallucinatory)} & \textbf{Model B: Llama-3.2-3B (Hallucinatory)} \\
\midrule
\textbf{0--4} & ``Ulm, Germany'', $p(C_1)=0.56$ & ``Ulm, Germany'', $p(C_1)=0.55$ \\
\textbf{5--7} & ``Baden-Württemberg, Germany'', $p(C_2)=0.30$ & ``Bern, Switzerland'', $p(C_2)=0.35$ \\
\textbf{8--9} & ``Germany'', $p(C_3)=0.14$ & ``Vienna, Austria'', $p(C_3)=0.10$ \\
\bottomrule
\end{tabular}
}
\end{table}

\paragraph{The Limitation of Semantic Entropy (SE)}
Both models produce three distinct semantic clusters with nearly identical probability distributions. SE solely computes the Shannon entropy of these cluster probabilities:
\begin{align*}
\text{SE}_A &= -0.56 \log_2(0.56) - 0.30 \log_2(0.30) - 0.14 \log_2(0.14) \approx 1.39 \text{ bits}, \\
\text{SE}_B &= -0.55 \log_2(0.55) - 0.35 \log_2(0.35) - 0.10 \log_2(0.10) \approx 1.34 \text{ bits}.
\end{align*}
Because SE ignores the semantic relationships between clusters, it incorrectly assigns similar uncertainty scores to both cases ($\Delta = 0.05$). It fails to recognize that Model A's answers are factually consistent (varying only in specificity), whereas Model B's answers are contradictory hallucinations.

\paragraph{SeSE and the Constructed Encoding Trees}
SeSE constructs a directed semantic graph using NLI entailment scores and builds an optimal $K=3$ encoding tree by minimizing structural entropy. As shown in Figure~\ref{fig:sese_worked_example}, the resulting trees reveal fundamentally different semantic organizations. 
By employing the optimal $3$-dimensional encoding tree $\mathcal{T}^{\star}$, SeSE captures the intrinsic hierarchical organization of the semantic space. The structural entropy of $\mathcal{T}^{*}$ quantifies the minimal number of bits required to describe this semantic space. Specifically, the information required to locate a leaf node in $\mathcal{T}^{A}$ is substantially less than that required for $\mathcal{T}^{B}$. When $K=1$, describing a leaf node's position in the one-dimensional encoding tree requires approximately $\log_2 10 \approx 3.32$ bits. When $K=3$, the hallucinatory case (Model B) exhibits a disordered semantic structure. It has higher uncertainty during random walks within the tree and greater resistance to compression, requiring $2.05$ bits to describe with a compression rate of $36\%$. By contrast, the non-hallucinatory case (Model A) possesses a more regular semantic organization. It is easier to compress and exhibits lower random-walk uncertainty, requiring only $1.60$ bits to describe with a compression rate of $46\%$. Consequently, SeSE effectively identifies fine-grained uncertainty distinctions even when existing semantic UQ methods fail. 



\section{Theoretical Proofs}
\label{appendix:theorem_proofs}
Semantic Entropy (SE) fundamentally assumes that the LLM's semantic space can be perfectly partitioned into disjoint semantic equivalence classes (i.e., semantic clusters). From a graph-theoretic perspective, this assumption is equivalent to compressing the fine-grained response-level graph we used into the coarse-grained \emph{quotient graph}. In this quotient graph, each semantic cluster acts as an atomic super-node, and the transition dynamics solely reflect the probability mass of the clusters, completely abstracting away the internal and inter-cluster topological structures. To formally prove that SeSE generalizes SE, we demonstrate that under SE's ideal conditions, applying SeSE to this quotient graph exactly recovers SE. 
In fact, in our practical implementation, we initialize with "clusters of size one"  (i.e, treating each singleton response as a cluster), whose transition probabilities are induced by NLI-based entailment scores between generated responses. This construction captures additional asymmetric and richer semantic dependencies beyond the coarse-grained cluster masses $\{p(C_i\mid x)\}$ used by SE. The detailed proof of Theorem~\ref{theorem:SeSE generalizes Semantic Entropy} is shown as follows.

\textbf{Theorem~\ref{theorem:SeSE generalizes Semantic Entropy} (SeSE Generalizes SE).} \textit{For any semantic clustering, there exists a semantic graph such that the one-dimensional structural entropy of this graph is equal to semantic entropy (computed as in Eq.~\ref{eq:se}\footnote{Here, we set logarithms to base 2 for the sake of clarity. Using a different base would only scale entropy values by a constant factor and does not affect relative uncertainty rank.}).} 

Let $C=\{C_1,\dots,C_M\}$ be an arbitrary semantic clustering, and let $p(C_i \mid x)$ denote the probability mass assigned to cluster $C_i$, which is estimated by the frequency of samples falling into each cluster. Therefore, we have $p(C_i \mid x)>0$ and $\sum_{i=1}^M p(C_i \mid x)=1$.

\begin{proof}
\textbf{Step 1: Construct a semantic quotient graph whose stationary distribution matches $p(C_i \mid x)_{i=1}^M$.}
We first construct a quotient graph $G_{\text{cluster}}$ where each vertex set represents a semantic cluster $C_i \in \{C_1,\dots,C_M\}$. Define a Markov transition matrix $P\in\mathbb{R}^{M\times M}$ by
\[
P(i,j) = p(C_j \mid x),\quad \forall\, i,j\in\{1,\dots,M\}.
\]
Since $\sum_{j=1}^M P(i,j)= \sum_{j=1}^M p(C_j \mid x)=1$, $P$ is row-stochastic. Consider a distribution $\pi\in\mathbb{R}^M$ given by $\pi(i) = p(C_i \mid x)$. For any $j$, 
\[
  (\pi P)(j) = \sum_{i=1}^M \pi(i) P(i,j) 
  = \sum_{i=1}^M p(C_i \mid x) p(C_j \mid x) 
  = p(C_j \mid x) \underbrace{\sum_{i=1}^M p(C_i \mid x)}_{=1} 
  = p(C_j \mid x) = \pi(j).
\]
Hence $\pi$ is a stationary distribution of $P$. Moreover, since $P(i,j)=p(C_j \mid x)>0$ for all $i,j$, the Markov chain is irreducible and aperiodic, so the stationary distribution $\pi$ is unique.

\textbf{Step 2: Apply SeSE with a one-dimensional encoding tree ($K=1$).}
Consider the one-dimensional encoding tree \(\mathcal{T}^1\) for \(G_{\text{cluster}}\), where the root \(\lambda\) contains all clusters, and each cluster \(C_i\) is represented by a distinct leaf node \(\alpha_i\) that is directly connected to the root, such that \(\mathcal{T}^1_{\alpha_i} = \{C_i\}\). By definition in the main text (Eq.~\ref{eq:one_directed_se}), the structural entropy of this single-layer encoding tree $\mathcal{T}^1$ is equivalent to the Shannon entropy of the stationary distribution of $G_{\text{cluster}}$:
\[
  H^{\mathcal{T}^1}(G_{\text{cluster}})=H^1(G_{\text{cluster}}) =-\sum_{i=1}^M \pi(i) \log_2 \pi(i).
\]
When the hierarchy height is restricted to $K=1$, the optimal encoding tree $\mathcal{T}^\star$ is uniquely determined as $\mathcal{T}^1$. As SeSE is defined as the total entropy of the optimal $K$-dimensional encoding tree $\mathcal{T}^*$, we therefore have
\[
  \mathrm{SeSE}(G_{\text{cluster}}, K=1) 
  = H^{\mathcal{T}^\star}(G_{\text{cluster}})
  = H^{\mathcal{T}^1}(G_{\text{cluster}})
  = -\sum_{i=1}^M \pi(i) \log_2 \pi(i).
\]
Substituting $\pi(i) = p(C_i \mid x)$, we obtain:
\[
  \mathrm{SeSE}(G_{\text{cluster}}, K=1)
  = -\sum_{i=1}^M p(C_i \mid x) \log_2 p(C_i \mid x)
  = \mathrm{SE}(x).
\]
Thus, we have proven that for any semantic clustering $C=\{C_1,\dots,C_M\}$ with distribution $\{p(C_i \mid x)\}_{i=1}^M$, there exists a corresponding semantic quotient graph such that SeSE with $K=1$ equals SE. 
\end{proof}

Theorem~\ref{theorem:SeSE generalizes Semantic Entropy} shows that SeSE not only recovers SE for any clustering, but is also more expressive than SE. When $K>1$, SeSE can capture the hierarchical structure of the semantic space through a multi-level encoding tree. The resulting optimal tree $\mathcal{T}^*$ reflects a progressive semantic partitioning: lower layers correspond to fine-grained partitions that capture subtle distinctions, while higher layers represent broader aggregations that reveal global semantic structural patterns. This hierarchical structure captures both local and global semantic relationships, enhancing the ability to distinguish subtle uncertainty differences. 
Consequently, SeSE can effectively distinguish uncertainties in complex scenarios where existing methods ~\cite{farquhar2024detecting, nikitin2024kernel, LiSYJCCR25,QiuM24} fail to differentiate between superficially similar semantic distributions.

\section{Details of SeSE in Long-form Generation}\label{appendix:long-form generation}
\begin{figure*}[htbp]
\centering
\includegraphics[width=\linewidth,trim={0cm 0 0cm 0}, clip]{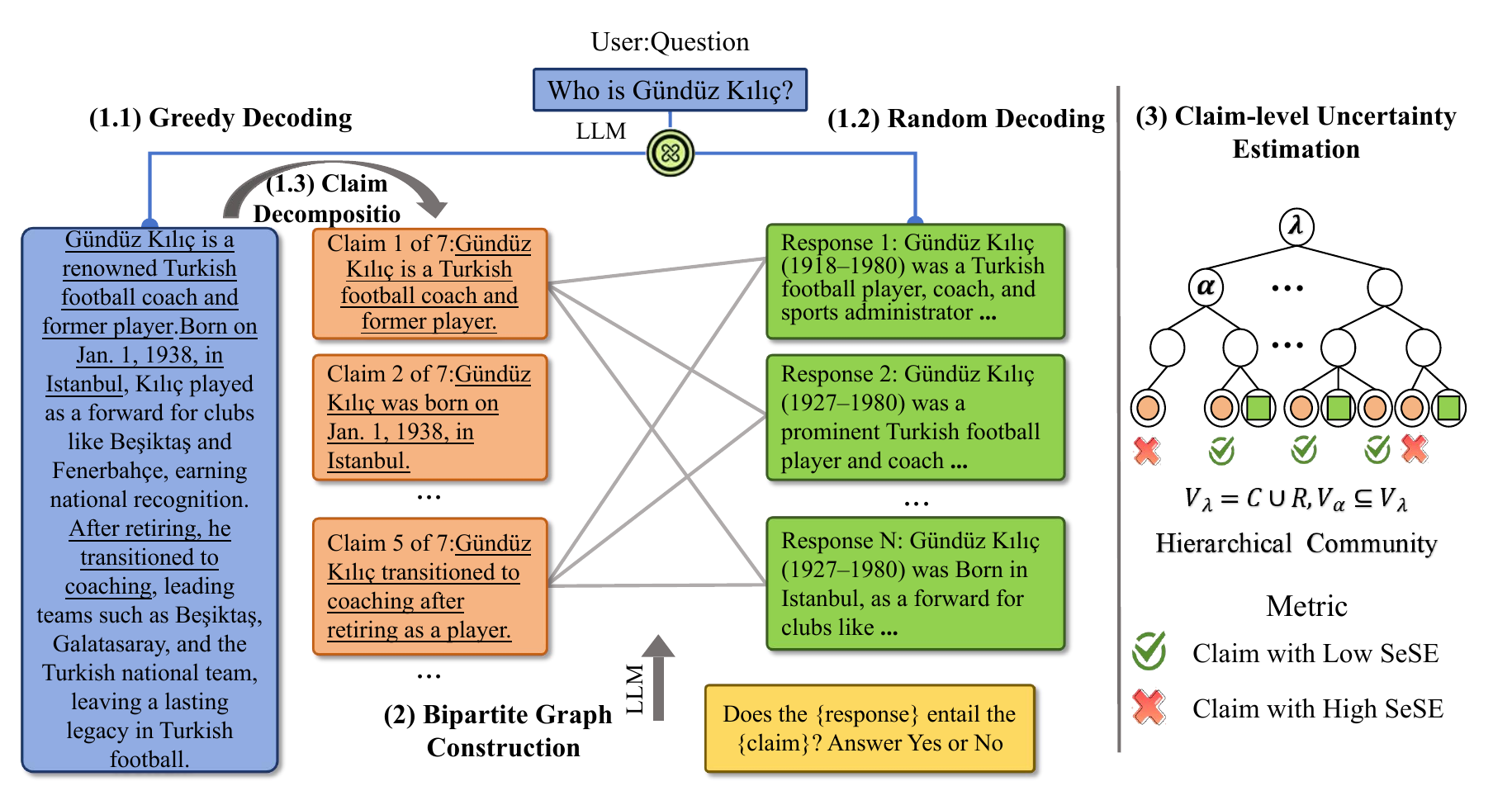}
\caption{Overview of SeSE in long-form generation. 
We decompose the generated long-form response into atomic claims. SeSE considers more sophisticated semantic structural information from the perspective of random walk, and hallucinations are indicated by high SeSE associated with that claim in the constructed bipartite response-claim graph.
}
\label{fig:SeSE-longform}
\end{figure*}

In practice, LLMs often output long-form paragraphs containing multiple \textbf{claims} \cite{min2023factscore}: the smallest semantically distinct unit of information presented within the generations. In long-form generation, we therefore assess uncertainty at the finer-grained claim level rather than simply assigning a single uncertainty score to an entire response or sentence.
We have the following observation: given a context $x$, a set of randomly sampled responses $R$, and a set of claims $C$ extracted from the greedily decoded response $r_{T=0}$, we can construct a bipartite graph $G_{cr} = ((R, C), E)$ where the edge set $E$ represents semantic entailment relationships between $R$ and $C$. 
The graph $G_{cr}$ thus captures semantic dependencies between responses $R$ and claims $C$, from which we can extract information that reflects the uncertainty associated with each claim $c \in C$. 
Intuitively, a claim that is densely connected to the response subgraph (i.e., consistently supported across sampled outputs) is more likely to be factually correct. Conversely, a sparsely connected claim is more likely to be a hallucination.
As shown in Fig.~\ref{fig:SeSE-longform}, we first construct $G_{cr} = ((R, C), E)$ and then estimate claim-level uncertainty using SeSE on this graph for hallucination detection.

\paragraph{Response Sampling and Claim Decomposition}
Using the same sampling settings as in Section~\ref{sec:SeSE for Short-form Generation}, we prompt LLMs to sample a stochastic response set $R(\cdot \mid x)$ and a greedily decoded response $r_{T=0}$.
Then, we prompt GPT-5-mini with a specific template to decompose $r_{T=0}$ into atomic claims, resulting in the claim set $C$. We adapt the prompt from \cite{jiang2024graph}. The prompt is as follows:
\begin{itemize}[leftmargin=\parindent, labelindent=0pt]
\item[]\textit{You will be provided with a long-form text that contains multiple claims. A claim is the smallest independent and self-contained perspective. Your task is to precisely identify and extract each claim within the given text, making sure there is no semantic repetition. Then, for the sake of clarity, resolve all anaphora (pronouns or other referring expressions) within the claims. Each claim should be concise and independently complete. Ensure that you are comprehensive and list each claim as a separate sentence.  \\
The input is:\{\textcolor{blue}{greedily decoded response}\} \\
Output:}
\end{itemize}

\paragraph{Bipartite Graph Construction}
The bipartite graph $G_{cr}$ is constructed by establishing connections between the response set $R$ and the claim set $C$. 
An edge $e \in E$ is created between a response $r \in R$ and a claim $c \in C$ if $r$ entails $c$. 
Edge weights are binary: $1$ if entailment holds and $0$ otherwise. 
We leverage the nuanced logical understanding of GPT-5-mini to assess entailment, thus avoiding the brittleness of manually tuned thresholds based on embedding distance or density-based clustering. Specifically, we adapt the following prompt from \cite{manakul2023selfcheckgpt} to each pair of response $r \in {R}$ and claim $c \in {C}$ to construct the edge set ${E}$ of the bipartite graph. Although this step requires $N * |C|$ LLM calls, it could take very little time (about 2 seconds per query using GPT-5-Mini) by \textbf{running in parallel rather than sequentially}.
\begin{itemize}[leftmargin=\parindent, labelindent=0pt]
\item[]\textit{Context: \{\textcolor{blue}{random sampling response}\}\\
Claim: \{\textcolor{blue}{claim}\}\\
Is the claim supported by the context above?\\
Answer Yes or No: \\
Output:}
\end{itemize}

And we presents the related ablation results of the entailment estimator in \textcolor{black}{Appendix}~\ref{sec:assessing entailment estimator}.

\paragraph{Claim-level Uncertainty Estimation}
In the bipartite graph $G_{cr}=((R, C), E)$, we model entailment relations as random walks between response and claim vertices, and quantify the uncertainty of these interactions using structural entropy. 
By minimizing the $K$-dimensional structural entropy of $G_{cr}$, we obtain its optimal encoding tree $\mathcal{T}^*_{cr}$, which captures the inherent hierarchical community structure over $C$ and $R$. Following the same process in Section~\ref{sec:SeSE for Short-form Generation}, we start by initializing a single-layer encoding tree $\mathcal{T}_{cr}$ in which each leaf node $\gamma$ has the tree root $\lambda$ as its parent. Then, we obtain the optimal $K-$dimensional encoding tree $\mathcal{T}^*_{cr}$ using Algorithm~\ref{alg:hierarchical_abstracting}.
In $\mathcal{T}^*_{cr}$, the root node $\lambda$ corresponds to the union of claim and response sets, $\mathcal{T}_{\lambda}=R \cup C$. Each leaf node $\gamma$ is a singleton containing an individual claim or response, and intermediate nodes represent hierarchical abstractions at different levels.

The structural entropy associated with each non-root node $\alpha$ quantifies the uncertainty of a random walk transitioning from the parent community $\mathcal{T}_{\alpha^-}$ to its child community $\mathcal{T}_{\alpha}$. For any claim $c \in C$, the uncertainty of reaching $c$ is determined by the cumulative entropy of all nodes $\alpha$ encountered along the path from the root node $\lambda$ to the leaf node $\gamma$ with $V_{\gamma} = \{c\}$. Accordingly, we define the SeSE of each claim $c$ as its uncertainty of engaging in random interactions within $G_{cr}$ as detailed below:
\begin{equation}
\text{SeSE}\left(G_{cr}; c\right) = -\sum_{\alpha \in \mathcal{P}(\lambda \to \gamma) \setminus \{\lambda\}} \frac{g_{\alpha}}{\operatorname{vol}(G_{cr})} \log_{2} \frac{\mathcal{V}_{\alpha}}{\mathcal{V}_{\alpha^{-}}},
\end{equation}

Nodes with low SeSE typically reside in the network's core regions, corresponding to claims that are frequently accessed during LLM generation, and are therefore more likely to be true.
Conversely, claims with high SeSE often occupy peripheral or sparsely connected regions, indicating a high likelihood of being hallucinations.

\section{Details of Hierarchical Abstraction}\label{appendix:Hierarchical Abstraction}
\paragraph{Merging and Combining Operators \cite{li2024science}}
\begin{figure*}[htbp]
	\centering
	\includegraphics[width=\linewidth]{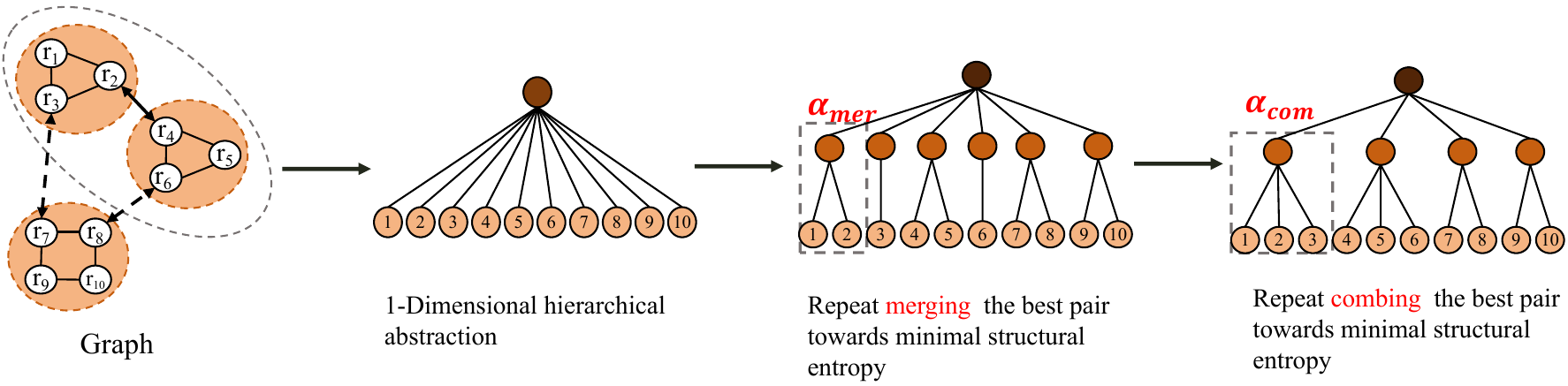}
	\caption{Illustration of the hierarchical abstraction construction with merging and combining operators.}
	\label{fig:hierarchical Abstraction}
\end{figure*}
Figure~\ref{fig:hierarchical Abstraction} illustrates the construction of a $2$-dimensional hierarchical abstraction. 
Specifically, consider a semantic space represented by a strongly connected graph $G=(V,W,E)$ with non-negative normalized edge weights, which is equal to an irreducible non-negative matrix $A_{n \times n}$. Let $V = \{1, 2, \cdots, n\}$, and let $\mathcal{T}$ be an encoding tree for $A$. Assume $\alpha$ and $\beta$ are two leaf nodes in $\mathcal{T}$ that share a common parent node $\gamma$, i.e., $\alpha^- = \beta^- = \gamma$. The steps for optimizing the encoding tree using the \textbf{merging} operator of leaf nodes $\alpha$ and $\beta$ are as follows:
\begin{enumerate}[label=(\arabic*)]
    \item Create a new node $\delta = \gamma^{\langle i \rangle}$, and renumber the child nodes of $\gamma$ as $0, 1, \cdots, k$;
    \item Set $T_\delta = \{x, y\}$;
    \item Create two new nodes $\delta^{\langle 0 \rangle},\delta^{\langle 1 \rangle}$;
    \item Define $T_{\delta^{\langle 0 \rangle}} = \{x\}$ and $T_{\delta^{\langle 1 \rangle}} = \{y\}$;
    \item Delete $\alpha$ and $\beta$.
\end{enumerate}

We define $g_{\alpha, \beta}$ as the total weight of edges connecting vertices in $V_{\alpha}$ to vertices in $V_{\beta}$ as follows:
\begin{equation}
    g_{\alpha, \beta} = \sum_{v_i \in V_{\alpha}} \sum_{v_j \in V_{\beta}} \pi(v_i) \cdot W(v_i, v_j).
\end{equation}
The entropy variation caused by a single merge operation on sibling nodes $\alpha, \beta \in T$ is denoted as $\Delta^{op_{mer}}_{\alpha, \beta}$ and is calculated as follows:
\begin{equation}
\resizebox{\textwidth}{!}{
$
\begin{aligned} 
\Delta^{op_{mer}}_{\alpha, \beta} 
&= \left[H^{\mathcal{T}}\left(G ; \alpha\right)+H^{\mathcal{T}}\left(G ; \beta\right)+\sum_{i}^{L_{\alpha}} H^{\mathcal{T}}\left(G ; \alpha_{i}\right)+\sum_{i}^{L_{\beta}} H^{\mathcal{T}}\left(G ; \beta_{i}\right)\right]-
\left[H^{\mathcal{T}'}\left(G ; \mu_{mer}\right) + \sum_{i}^{L_{\alpha}} H^{\mathcal{T}'}\left(G ; \alpha_{i}\right)+\sum_{i}^{L_{\beta}} H^{\mathcal{T}'}\left(G ; \beta_{i}\right)\right] \\
& =\frac{g_{\alpha}-\sum_{i}^{L_{\alpha}} g_{\alpha_{i}}}{\operatorname{vol}\left(G\right)} \cdot \log _{2} \frac{\mathcal{V}_{\mu_{mer}}}{\mathcal{V}_{\alpha}}+\frac{g_{\beta}-\sum_{i}^{L_{\beta}} g_{\beta_{i}}}{\operatorname{vol}\left(G\right)} \cdot \log _{2} \frac{\mathcal{V}_{\mu_{mer}}}{\mathcal{V}_{\beta}}+\frac{g_{\alpha, \beta}+g_{\beta, \alpha}}{\operatorname{vol}\left(G\right)} \cdot \log _{2} \frac{\mathcal{V}_{\alpha^{-}}}{\mathcal{V}_{\mu_{mer}}} \\ & =\frac{g_{\alpha, \beta}+g_{\beta, \alpha}}{\operatorname{vol}\left(G\right)} \cdot \log _{2} \frac{\mathcal{V}_{\alpha^{-}}}{\mathcal{V}_{\mu_{mer}}}-\frac{\sum_{i \neq j}^{L_{\alpha}} g_{\alpha_{i}, \alpha_{j}}}{\operatorname{vol}\left(G\right)} \cdot \log _{2} \frac{\mathcal{V}_{\mu_{mer}}}{\mathcal{V}_{\alpha}}-\frac{\sum_{i \neq j}^{L_{\beta}} g_{\beta_{i}, \beta_{j}}}{\operatorname{vol}\left(G\right)} \cdot \log _{2} \frac{\mathcal{V}_{\mu_{mer}}}{\mathcal{V}_{\beta}},
\end{aligned}
$
}
\end{equation}
where $\mathcal{T}'$ represents the tree after the merge operation, $L_\alpha$ denotes the number of child nodes of $\alpha$, and $\mu_{mer}$ is the newly added node created by the merge operation.

Assume $\alpha$ and $\beta$ are two arbitrary nodes in $\mathcal{T}$ that share a common parent node $\gamma$, i.e., $\alpha^- = \beta^- = \gamma$. The steps for optimizing the encoding tree using the \textbf{combining} operator of $\alpha$ and $\beta$ are as follows:
\begin{enumerate}[label=(\arabic*)]
    \item Let $T_\alpha$ and $T_\beta$ denote the subtrees of $\mathcal{T}$ rooted at $\alpha$ and $\beta$, respectively;
    \item Create a new node $\delta$ with parent $\gamma$ (i.e., $\delta$ shares the same parent as $\alpha$ and $\beta$);
    \item Add two child nodes to $\delta$: $\delta^{\langle 0 \rangle}$ and $\delta^{\langle 1 \rangle}$;
    \item Insert the subtree $T_\alpha$ and $T_\beta$ into $\delta^{\langle 0 \rangle}$ and $\delta^{\langle 1 \rangle}$, respectively;
    \item Delete $\alpha$ and $\beta$.
\end{enumerate}
The entropy variation caused by a single combine on sibling nodes $\alpha, \beta \in T$ is denoted as $\Delta^{op_{com}}_{\alpha, \beta}$ and is calculated as follows:
\begin{equation}
\begin{aligned} 
\Delta^{op_{com}}_{\alpha, \beta} & =\left[H^{\mathcal{T}}\left(G ; \alpha\right)+H^{\mathcal{T}}\left(G ; \beta\right)\right]-
\left[H^{\mathcal{T}'}\left(G ; \mu_{com}\right)+H^{\mathcal{T}'}\left(G ; \alpha\right)+H^{\mathcal{T}'}\left(G ; \beta\right)\right] \\ 
& =\frac{g_{\alpha}}{\operatorname{vol}\left(G\right)} \cdot \log _{2} \frac{\mathcal{V}_{\alpha^{-}}}{\mathcal{V}_{\mu_{com}}}+\frac{g_{\beta}}{\operatorname{vol}\left(G\right)} \cdot \log _{2} \frac{\mathcal{V}_{\alpha^{-}}}{\mathcal{V}_{\mu_{com}}}-\frac{g_{\mu_{com}}}{\operatorname{vol}\left(G\right)} \cdot \log _{2} \frac{\mathcal{V}_{\alpha^{-}}}{\mathcal{V}_{\mu_{com}}} \\ & =\frac{g_{\alpha}+g_{\beta}-g_{\mu_{com}}}{\operatorname{vol}\left(G\right)} \cdot \log _{2} \frac{\mathcal{V}_{\alpha^{-}}}{\mathcal{V}_{\mu_{com}}} \\ & =\frac{g_{\alpha, \beta}+g_{\beta, \alpha}}{\operatorname{vol}\left(G\right)} \cdot \log _{2} \frac{\mathcal{V}_{\alpha^{-}}}{\mathcal{V}_{\mu_{com}}} ,
\end{aligned}
\end{equation}
where $\mu_{com}$ is the newly added node via the combine operation.
The $2$-dimensional encoding tree could then be optimized to the needed $K$-dimension by continuing to greedily and iteratively apply merging and combining operators.

\paragraph{Time Complexity of Hierarchical Abstraction Construction}
The overall time complexity of the hierarchical abstraction construction is $O(n^2 + m \cdot \log_2 n)$ (Step~2 and Step~3 in subsection~\ref{sec:SeSE for Short-form Generation}), where $n = |V|$ denotes the number of vertices in the semantic graph, and $m = |E|$ indicates the number of edges. The graph construction phase exhibits a time complexity of $O(n^2)$. According to the analysis of \cite{paninformation}, the optimization process of the high-dimensional encoding tree via merging and combining operators contributes a time complexity of $O(m \cdot \log_2 n)$.

\section{Additional Experimental Details and Analysis}\label{appendix:Additional Experimental Details}

\subsection{Hardware and Resources}
In terms of computing resources, as it is necessary to sample generations from LLMs to model the semantic space, our experiments require one or more GPUs to accelerate LLMs inference. Without GPU support, reproducing the results within a reasonable timeframe is infeasible. For short-form generation tasks, we use the GPT-5-mini model accessed through the OpenAI API to evaluate accuracy. As OpenAI's pricing is based on the number of input and output tokens, the cost of reproducing our experiments varies with configuration, typically averaging around 1\$-5\$ per run.
The concurrent experiments are conducted on two NVIDIA RTX PRO 6000 graphics cards, each with 96 GB of memory. Depending on the model size and experimental setup, the generation phase for each scenario requires between 2 and 24 hours. Our experimental procedure involves first generating responses for all dataset-model pairs, followed by the computation of various uncertainty metrics. Model outputs are not regenerated across runs; instead, only the corresponding uncertainty metrics are recalculated.
\subsection{Comparison of Computational Resource Consumption}\label{sec:compare_resource}
In this subsection, we analyze the computational resource consumption of various UQ methods. Due to the large parameter sizes of LLMs, their inference costs are substantially higher than those of other components. Our analysis thus focuses primarily on LLM inference consumption.  
Besides the white-box method LN-PE, the few-shot prompting method P(True), and the supervised training method ER, all other baselines require sampling $N$ possible answers, incurring the same consumption. To improve
the accuracy in semantic clustering for SE and self-consistency assessment for SC, following the original implementations, we employ GPT-5-mini for entailment prediction. In the worst-case scenario, this necessitates additional $N^2$ LLM inference calls and $N$ calls for SC. Graph-based methods (KLE and SGD) utilize lightweight NLI models to compute entailment scores between responses, which is identical to SeSE.
In summary, SeSE significantly reduces computational costs compared to SE and SC while achieving superior performance. When compared to graph-based methods, SeSE delivers better results while maintaining equivalent resource consumption. Although the sampling process increases the generation cost, SeSE avoids the limitations of supervised methods (ER) that require retraining for new models and tasks, or white-box methods (LN-PE, P(True)) that depend on LLM internal states. Moreover, in safety-critical tasks, the potential cost of a hallucination should outweigh the cost of sampling multiple answers. Therefore, reliable uncertainty quantification through SeSE should always be worthwhile.

\subsection{Generalization Analysis}
\begin{figure}[htbp]
    \centering
    \begin{subfigure}[t]{0.49\linewidth}
        \centering
        \includegraphics[width=\linewidth]{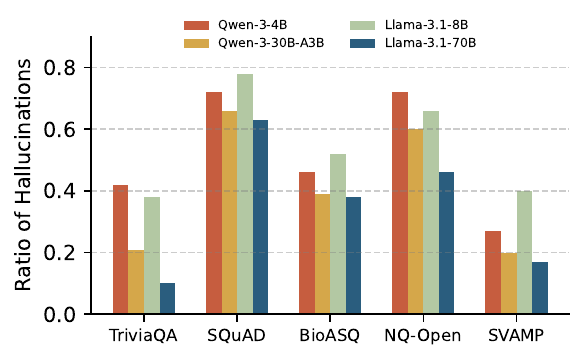}
        \caption{Hallucination rate of LLMs across used datasets.}
    \label{fig:llm_hallucination_datasetlevel}
    \end{subfigure}
    \begin{subfigure}[t]{0.49\linewidth}
        \centering
        \includegraphics[width=\linewidth]{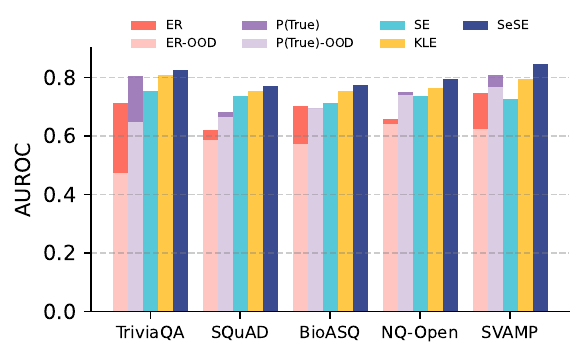}
        \caption{OOD performance comparison among entropy-based methods.}
    \label{fig:datalevel_generalization_auroc}
    \end{subfigure}
    \caption{Hallucination rate of used LLMs in different domains and performance comparison of entropy-based methods in OOD datasets. OOD represents the method is evaluated on out-of-distribution datasets.}
    \label{fig:llm_acc}
\end{figure}

Figure \ref{fig:llm_hallucination_datasetlevel} illustrates the hallucination rates of used LLMs on each dataset. As can be seen, our experiments cover a range of LLMs with varying hallucination levels across diverse generation tasks.
The values plotted in Figure \ref{fig:datalevel_generalization_auroc} represent the aggregate AUROC scores over five LLMs. Embedding regression is a representative supervised approach that utilizing a trained logistic regression classifier to predict answer correctness. P(True) serves as an ``in-context" supervised method that adapts to specific tasks through few-shot demonstrations in the prompt. 
As indicated by the light red and light purple bars, both P(True) and ER suffer substantial performance degradation when the data distribution shifts between training and testing. In contrast, as an ``off-the-shelf'' method, SeSE consistently outperforms both supervised and entropy-based baselines on in-distribution and out-of-distribution (OOD) datasets, demonstrating significant generalization and practical utility potential. Such generalization is especially important for UQ, as real-world scenarios frequently involve distribution shifts between training and deployment phases, and reliable UQ methods should perform well across different scenarios.

\begin{figure}[htbp]
	\centering
	\includegraphics[width=0.7\linewidth]{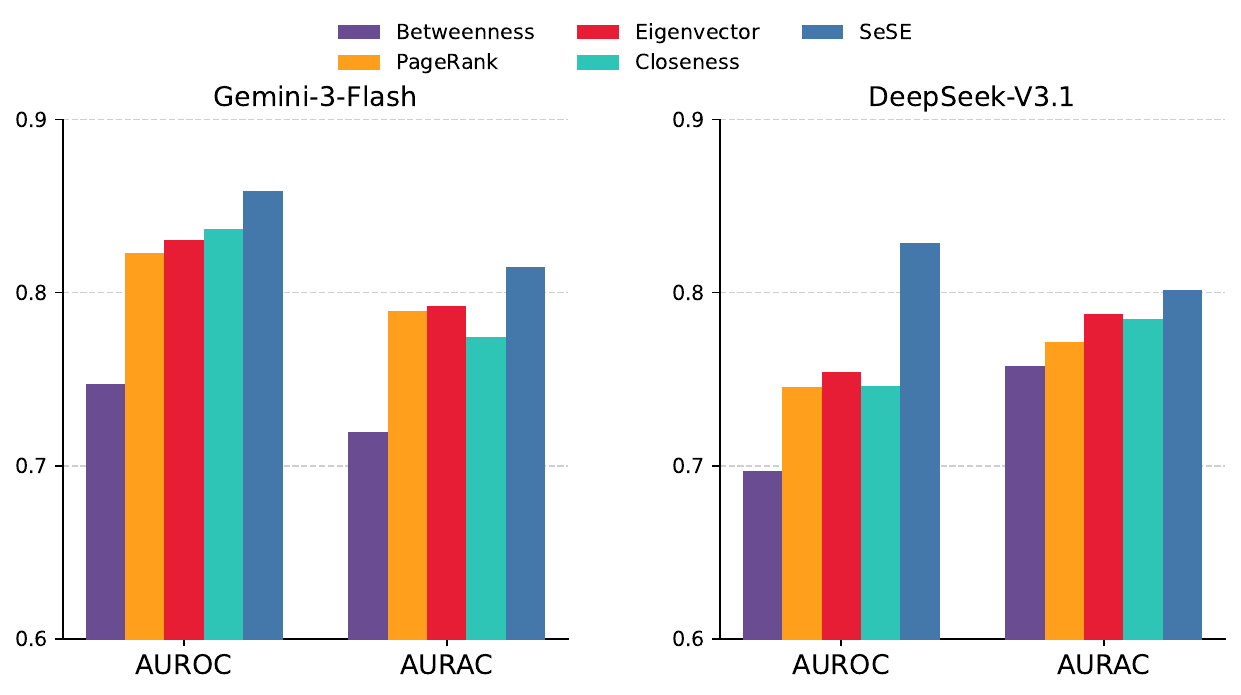}
	\caption{SeSE significantly outperforms benchmark graph centrality metrics in Table \ref{tab:graph_centrality_metrics}.}
	\label{fig:graph centrality metric}
\end{figure}

\subsection{Comparison with Graph Centrality Metrics}
Regarding granular uncertainty estimation in long-form generation, the most closely related work to ours is Graph Uncertainty \cite{jiang2024graph}, which utilizes the negative centrality scores of claim nodes within a claim-response bipartite graph as the uncertainty metric. However, the centrality metrics employed by \citep{jiang2024graph} fail to capture the richer semantic topological structures \cite{li2016structural} and offer limited interpretability. Here, we compare SeSE with several widely used graph centrality measures, including betweenness, eigenvector, PageRank, and closeness \cite{lingenerating}. The specific definitions of these metrics are detailed in Table \ref{tab:graph_centrality_metrics}. We follow the same setup in main experiments. As illustrated in Figure \ref{fig:graph centrality metric}, SeSE consistently outperforms these graph centrality metrics, demonstrating its superior ability to identify central nodes within the claim-response graph.

\begin{table*}[htbp]
\centering
\renewcommand{\arraystretch}{1} 
\caption{
Graph centrality metrics with their formulas and explanations. 
}
\begin{threeparttable}
\resizebox{\textwidth}{!}{
\begin{tabular}{@{}lll@{}}
\toprule[1pt]
\textbf{Metric} & \textbf{Formula} & \textbf{Brief Explanation} \\
\midrule[0.75pt]
\multirow{2}{*}{Betweenness} & \multirow{2}{*}{$C_B(v) = \sum_{s \neq v \neq t} \frac{\sigma_{st}(v)}{\sigma_{st}}$} & \multirow{2}{8cm}{Fraction of shortest paths $\sigma_{st}$ between other nodes $s$, $\mathcal{T}$ that pass through a node $v$.} \\ \\
\cmidrule{1-3}
\multirow{3}{*}{Eigenvector} & \multirow{3}{*}{$C_{Eigv}(v) = \frac{1}{\lambda} \sum_{u \in N(v)} A_{vu} C_{Eigv}(u)$} & \multirow{3}{8cm}{Evaluates the influence of node $v$ based on the importance of its neighbors $N(v)$. \( A_{vu} \) is adjacency matrix entry. \( \lambda \) is the largest eigenvalue of $A$.} \\ \\ \\ 
\cmidrule{1-3}
\multirow{3}{*}{PageRank} & \multirow{3}{*}{$C_{PR}(v) = \frac{1-d}{|V|} + d \sum_{u \in N(v)} \frac{C_{PR}(u)}{|N(u)|}$} & \multirow{3}{8cm}{Quantifies node importance by combining link quantity and quality. $d$ is the damping factor. $N(v)$ is the set of neighboring nodes of node $v$.} \\ \\ \\
\cmidrule{1-3}
\multirow{3}{*}{Closeness} & \multirow{3}{*}{$C_C(v) = \frac{|V|-1}{\sum_{u \in V} d(v,u)} \cdot \frac{|V|}{|V_v|}$} & \multirow{3}{8cm}{Reciprocal of the average shortest path distance to all nodes. \( d(v, u) \) is the shortest-path distance between \( v \) and \( u \). \( |V_v| \) is number of nodes reachable from \( v \).} \\ \\ \\
\bottomrule[1pt]
\end{tabular}
}
\begin{tablenotes}
\item $V$: The node set of graph $G$. $A$: The adjacency matrix of graph $G$. $|V|$: The total number of nodes in graph $G$.
\end{tablenotes}
\end{threeparttable}
\label{tab:graph_centrality_metrics}
\end{table*}

\subsection{Assessing Accuracy of Automated 
Ground-truth Evaluations}\label{sec:automated 
ground-truth evaluations}
The F1 score is the harmonic mean of precision and recall of the lexical overlap between the reference answer and the generated answer. 
It is widely used to evaluate fixed-answer generation tasks.
However, this metric exhibits obvious limitations in short-form, free-form text generation: lexical overlap between LLMs responses and the short reference answers may be unreasonably low, rendering the F1 score ineffective.
Therefore, our study leverages the natural language understanding capabilities of LLMs rather than relying on simple lexical matching. We employ GPT-5-mini to assess semantic equivalence between LLM-generated answers and reference answers.

To verify the reliability of our automated factual evaluation, we manually inspect 500 questions (100 from each of five experimental datasets) and analyze the short-form answers generated by the models. We focus on the concordance between human and automated methods, rather than the correctness of the evaluations. {Table~\ref{tab:Automated Factual Assessment}} presents consistency statistics between automated evaluation methods and human reviewer judgments. The table shows that the agreement rate between the two human assessors (95\%) closely approximates their average agreement rate with GPT-5-mini (94\%). Although GPT-4o and Qwen-3-32B perform slightly worse, we select GPT-5-mini for the results presented in this paper, as it provides the best factual estimates.

\begin{table*}[hptb]
\centering
\caption{Assessing automated ground-truth evaluators.}
\resizebox{0.8\textwidth}{!}{
\begin{tabular}{@{}lcccccc@{}}
\toprule
& F1 Score & Qwen-3-32B & GPT-4o & GPT-5-mini & Human A & Human B \\
\midrule
Human A & 0.63 & 0.93 & 0.93 & 0.95 & - & 0.95 \\
Human B & 0.60 & 0.91 & 0.91 & 0.93 & 0.95 & - \\
Average & 0.62 & 0.92 & 0.92 & 0.94 & - & - \\
\bottomrule
\end{tabular}
}
\label{tab:Automated Factual Assessment}
\end{table*}

\subsection{Assessing Entailment Estimator in Long-form Generation}\label{sec:assessing entailment estimator}
\begin{figure*}[htbp]
	\centering
	\includegraphics[width=0.8\linewidth]{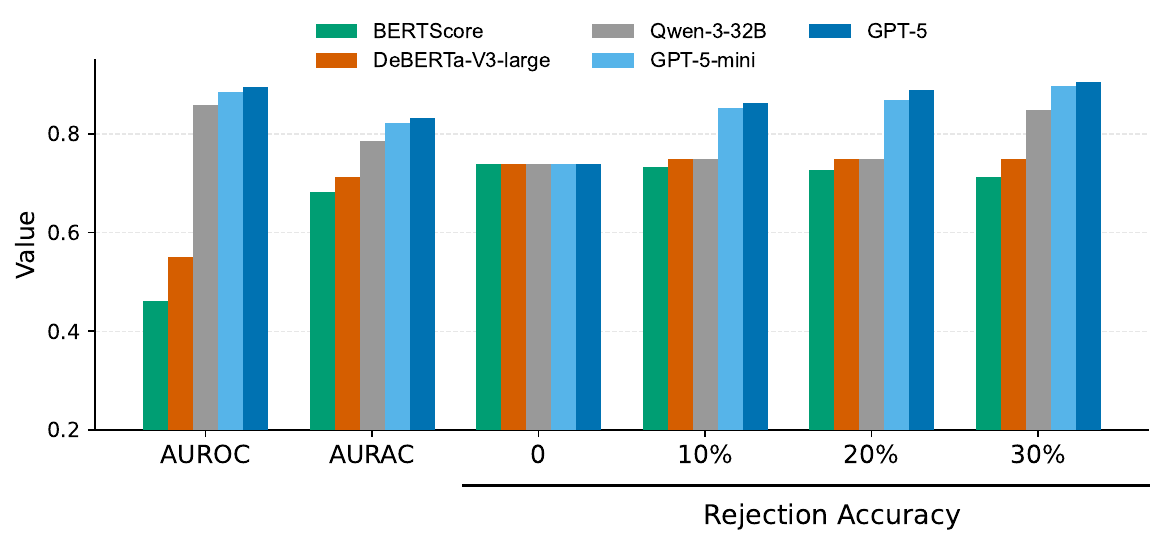}
	\caption{Assessing entailment estimators in long-form generation. The 10\% rejection accuracy indicates the accuracy of the LLM after declining to respond to queries whose uncertainty ranking falls within the top 10\%.}
	\label{fig:long-form entailment estimator}
\end{figure*}

The bipartite graph construction involves entailment judgments between long-form responses and short claims. Figure~\ref{fig:long-form entailment estimator} shows the ablation results of different entailment estimators. Conventional BERTScore and DeBERTa perform poorly, and GPT-5 does not offer notable advantages compared to GPT-5-mini. Consequently, we choose GPT-5-mini for its comparable performance and greater cost-effectiveness. The experiment is conducted on the PopQA using Gemini-3-Flash.

\begin{figure*}[htbp]
    \centering
    \begin{subfigure}[t]{0.49\linewidth}
        \centering
        \includegraphics[width=\linewidth]{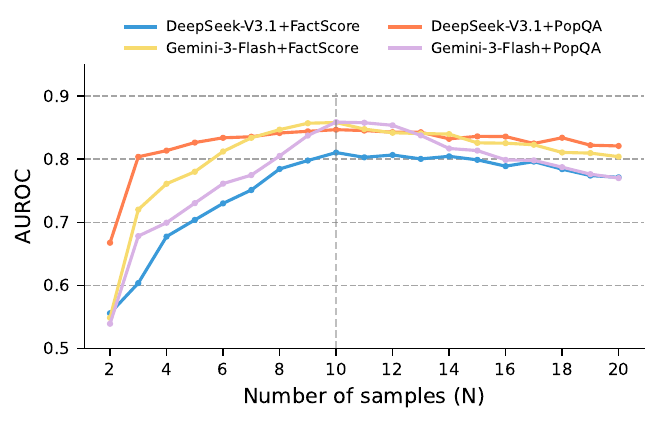}
    \end{subfigure}
    \hfill
    \begin{subfigure}[t]{0.49\linewidth}
        \centering
        \includegraphics[width=\linewidth]{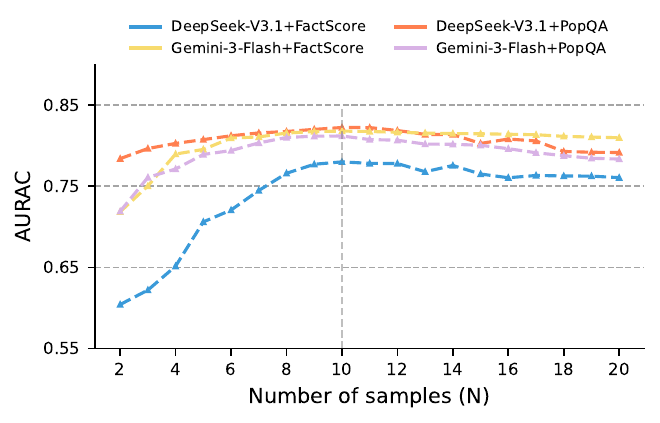}
    \end{subfigure}
    \caption{AUROC and AURAC performance across different sample sizes in long-form experiments.} 
    \label{fig:ungrated sensitivity to N}
\end{figure*}

\begin{figure*}[htbp]
    \centering

    \begin{subfigure}[t]{\textwidth}  
        \centering
        \includegraphics[width=0.9\textwidth]{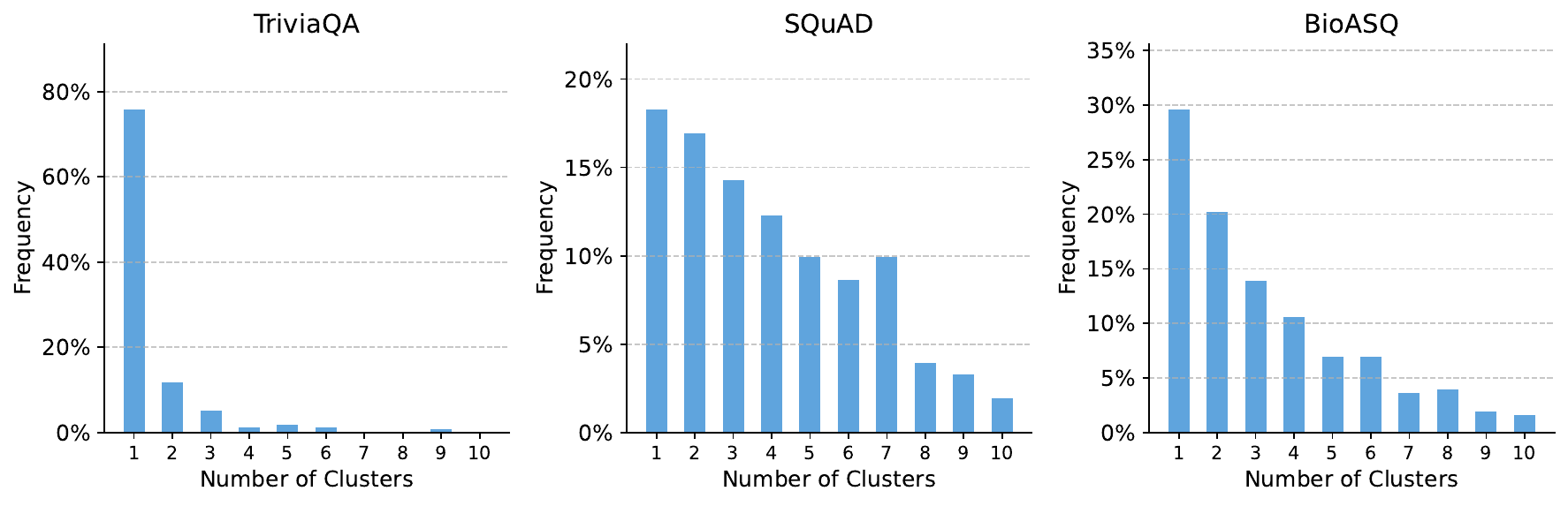}
    \end{subfigure}
   
    \begin{subfigure}[t]{0.66\textwidth}  
        \centering
        \includegraphics[width=\textwidth]{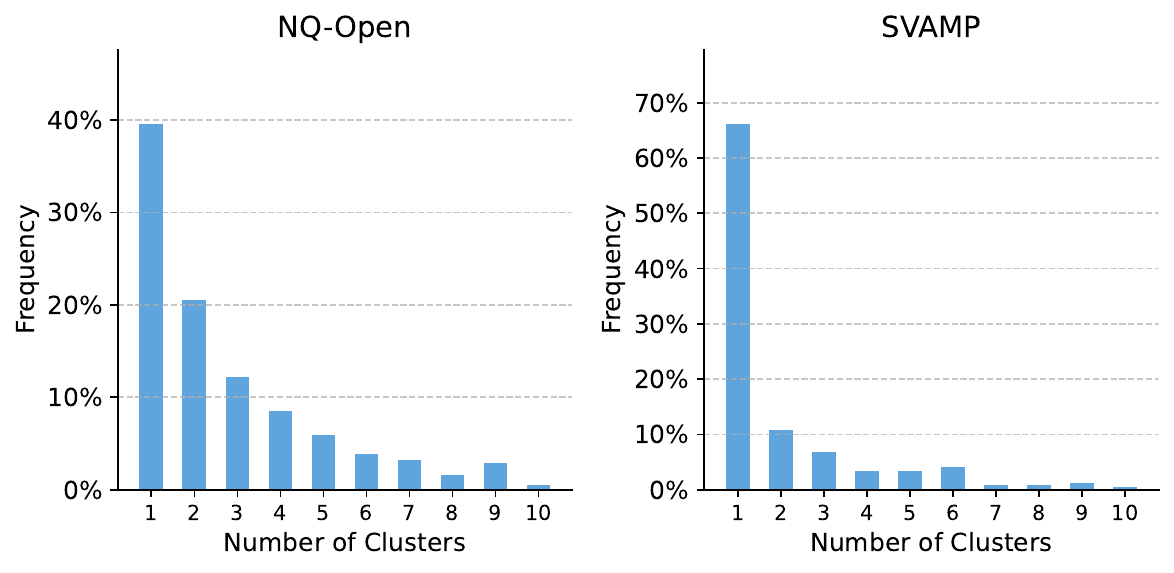}
    \end{subfigure}
    
    \caption{Statistics of semantic cluster numbers across datasets used in the short-form experiment. All plots are based on generations of Llama-3.1-70B.}
    \label{fig:cluster_num_across_datasets}
\end{figure*}

\begin{figure*}[htbp]
	\centering
	\includegraphics[width=\linewidth]{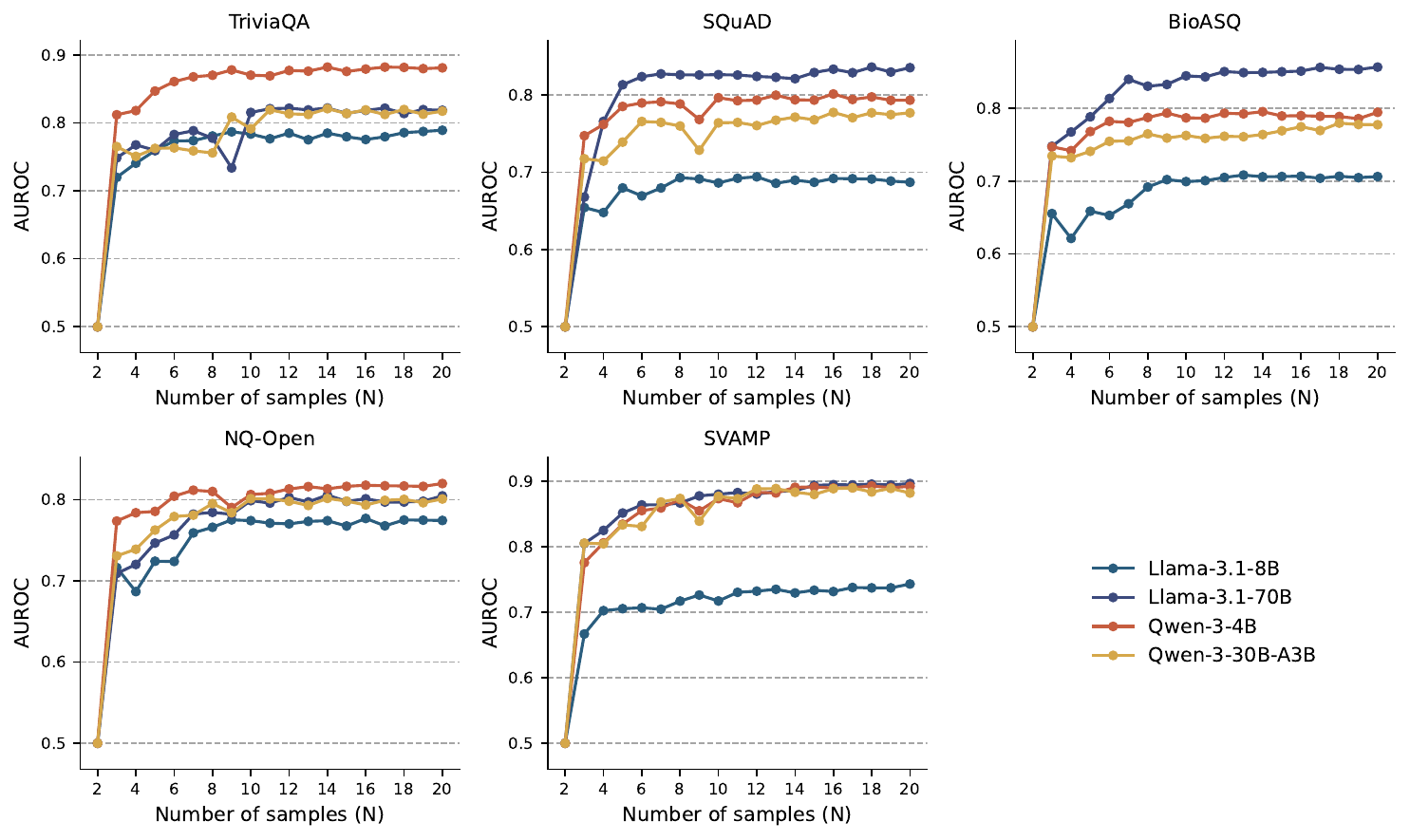}
	\caption{AUROC performance across different sample sizes in short-form experiments.}
 \label{fig:detailed_sentence_length_N_auroc}
\end{figure*}

\begin{figure*}[htbp]
	\centering
	\includegraphics[width=\linewidth]{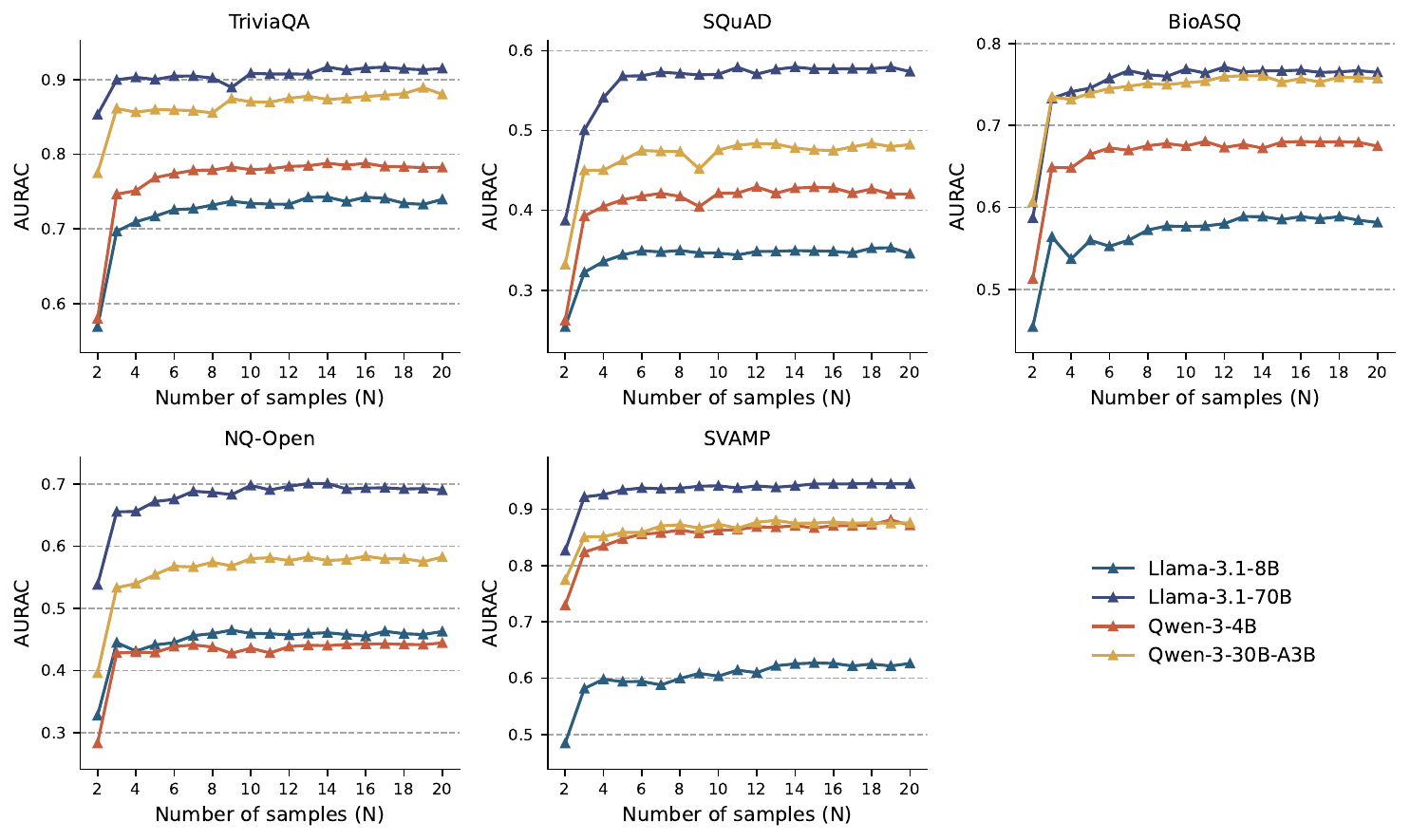}
	\caption{AURAC performance across different sample sizes in short-form experiments.}
 \label{fig:detailed_sentence_length_N_aurac}
\end{figure*}

\subsection{Detailed Results of Hyperparameter Sensitivity}\label{sec:Detailed results of hyperparameter sensitivity}
In Figure \ref{fig:cluster_num_across_datasets}, we present the empirical statistics of semantic cluster counts across datasets used in the short-form experiment. As illustrated, the semantic spaces of simpler datasets such as TriviaQA and SVAMP usually contain only one or two clusters, reflecting the fact that large language models generally answer these questions correctly. In contrast, more challenging datasets like SQuAD exhibit much higher semantic complexity: approximately 30\% of the questions correspond to semantic spaces with six or more clusters, indicating more intricate and disordered semantic structures. From an information-theoretic perspective, such complex semantic spaces are difficult to compress. Accurately describing them therefore requires more information (bits), i.e., constructing deeper encoding trees (larger $K$) to effectively quantify their inherent uncertainty.
Figure~\ref{fig:ungrated sensitivity to N} shows the unaggregated hyperparameter sensitivity results in the long-form experiments.
In Figures~\ref{fig:detailed_sentence_length_N_auroc}---\ref{fig:detailed_sentence_length_N_aurac}, we report the unaggregated hyperparameter sensitivity results of the sampling size $N$ in short-form experiments.

\newpage
\section{Prompt Details}\label{appendix:Prompt}
Here we mark \textcolor{blue}{placeholders} with the blue color.

\subsection{Response Sampling Prompt}
We follow \cite{farquhar2024detecting} and use the following prompt template to obtain answers in short-form experiments, including both the greedily-decoded answer (to evaluate the model's accuracy) and the randomly sampled responses (to measure the model's uncertainty).
\begin{itemize}[leftmargin=\parindent, labelindent=0pt]
\item[]\textit{Answer the following question concisely in one sentence. \\
Question: \{\textcolor{blue}{question}\}\\
Answer:}
\end{itemize}

Table~\ref{tab:example_responses} includes two examples showing the sampled responses of the Llama-3.1-8B to two questions.

\begin{table*}[t]
\centering
\small
\caption{Example Responses from BioASQ and NQ-Open datasets sampled from Llama-3.1-8B.}
\label{tab:example_responses}
\begin{tabularx}{\textwidth}{XX}
\toprule
\textbf{Example Responses from BioASQ} & \textbf{Example Responses from NQ-Open} \\
\midrule
{\textbf{Input:} Answer the following question concisely in one sentence. Question: Which organs are mostly affected in Systemic Lupus Erythematosus (SLE)? Answer:} & 
{\textbf{Input:} Answer the following question concisely in one sentence. Question: Which financial statement involves all aspects of the accounting equation? Answer:}\\
\addlinespace[0.5em]
\textbf{Reference Answer} & \textbf{Reference Answer} \\
kidney, brain, heart, skin & balance sheet \\
\addlinespace[0.5em]
\textbf{Greedily-Decoded Response} & \textbf{Greedily-Decoded Response} \\
SLE primarily affects the skin, kidneys, and heart. \correct & accounting equation. \wrong \\
\midrule
\textbf{Randomly Sampled Responses} & \textbf{Randomly Sampled Responses} \\
\begin{itemize}[leftmargin=1.5em, nosep, labelsep=0.5em]
    \item $r^{(1)}$: including the skin, joints, kidneys, and kidneys
    \item $r^{(2)}$: the skin, joints, kidneys, and heart.
    \item $r^{(3)}$: the skin, joints, kidneys, heart, brain, and kidneys.
    \item $r^{(4)}$: the skin, joints, kidneys, and the immune system.
    \item $r^{(5)}$: the skin, joints, kidneys, and lymph nodes.
    \item $r^{(6)}$: the joints, skin, kidneys, and kidneys.
    \item $r^{(7)}$: the skin, kidneys, heart, brain, and immune system.
    \item $r^{(8)}$: the skin, kidneys, and lupus-prone joints.
    \item $r^{(9)}$: the skin, joints, kidneys, heart, and brain.
    \item $r^{(10)}$: the skin, kidneys, heart, and lungs.
\end{itemize} 
& 
\begin{itemize}[leftmargin=1.5em, nosep, labelsep=0.5em]
    \item $r^{(1)}$: Balance Sheet.
    \item $r^{(2)}$: the balance sheet.
    \item $r^{(3)}$: balance sheet.
    \item $r^{(4)}$: balance sheet.
    \item $r^{(5)}$: Income statement.
    \item $r^{(6)}$: the statement of cash.
    \item $r^{(7)}$: the cash flow statement.
    \item $r^{(8)}$: the Balance sheet.
    \item $r^{(9)}$: the statement of equity or stockholders' equity.
    \item $r^{(10)}$: the cash flow statement.
\end{itemize} \\
\bottomrule
\end{tabularx}
\end{table*}

\subsection{Ground-truth Evaluation Prompt}
For short-form generation, we automatically determine whether the given answer is correct or incorrect by comparing it with the reference answer using GPT-5-mini. The specific prompt that we adapted from \cite{farquhar2024detecting} is as following:
\begin{itemize}[leftmargin=\parindent, labelindent=0pt]
\item[] \textit{ We are evaluating the correctness of answers to the question: \{\textcolor{blue}{question}\} \\
The reference answer is: \{\textcolor{blue}{reference answer}\} \\
The proposed answer is: \{\textcolor{blue}{greedy decoding generation}\} \\
According to the reference answer, determine whether the proposed answer is correct within the context of the question. Respond with only Yes or No.}
\end{itemize}
For the dataset SQuAD\_V2 with multiple reference answers, the second line becomes ``The following are the reference answers:", and the last line asks ``determine whether the proposed answer has the same meaning as any of the reference answers within the context of the question."

\section{Datasets Details}\label{appendix:Dataset}
SeSE can detect confabulations in free-form text generation across a range of domains without requiring prior domain knowledge. We evaluate it on short-form question-answering tasks spanning life sciences, mathematical reasoning, trivia knowledge, open-domain QA, and commonsense reasoning. Furthermore, to examine SeSE's hallucination detection capability in long-text generation tasks, we construct four custom datasets based on the generated outputs of DeepSeek-V3.1 and Gemini-3-Flash on two benchmark datasets---FActScore and PopQA. Since current LLMs have an excessively high accuracy rate when using tools, all experiments are conducted in offline mode, relying solely on their own capabilities. All experimental datasets have been made publicly available in the code repository to facilitate reproducibility and further development.

\subsection{Datasets in Short-form Experiments}
In the short-form experiment, we utilize five representative QA datasets covering different domains: BioASQ \cite{krithara2023bioasq}, SVAMP \cite{patel2021nlp}, TriviaQA \cite{joshi2017triviaqa}, NQ-Open \cite{nqopen2019}, and SQuAD\_V2 \cite{rajpurkar-etal-2018-know}. 
BioASQ derives from the annual biomedical semantic-indexing and question-answering challenge of the same name, focusing on life sciences. We select dataset from Task B of the 2023 BioASQ challenge. 
SQuAD\_V2, a reading comprehension dataset, contains answers extracted from Wikipedia paragraphs. We exclude unanswerable questions designed to induce erroneous responses. 
TriviaQA encompasses trivia questions across multiple domains, including history, science, entertainment and so on. 
SVAMP features elementary mathematical word problems that test reasoning abilities.
NQ-Open, an open-domain subset of Natural Questions, comprises real Google search queries with answers found in Wikipedia documents.
For each dataset, we sample 300 examples for training and 300 for testing each time with different random seed. 
Notably, SeSE is completely independent of the training data. For datasets with predefined partitions, sampling occurred within their respective training and test sets. 
All experiments employ free-form question answering rather than multiple-choice or true/false formats to better assess the characteristics of LLMs in generating free-from text. As the questions become too easy for the current LLMs when context is provided, we withhold context paragraphs across all experiment datasets except SVAMP to increase difficulty. 

Below are some example prompts of the short-form datasets we used, which are feed to the LLMs according to the template described in Appendix~\ref{appendix:Prompt}.
\begin{quespromptbox}{BioASQ}
\textit{Answer the following question concisely in one sentence.}\\
\textbf{Question:} What is the msDNA?\\
\textbf{Answer:}\\
\textbf{Reference Answer:} msDNA is actually a complex of DNA, RNA, and probably protein.
\end{quespromptbox}

\begin{quespromptbox}{NQ-Open}
\textit{Answer the following question concisely in one sentence.}\\
\textbf{Question:} What is the name of india pakistan border?\\
\textbf{Answer:}\\
\textbf{Reference Answer:} International Border (IB)
\end{quespromptbox}

\begin{quespromptbox}{SQuAD\_V2}
\textit{Answer the following question concisely in one sentence.}\\
\textbf{Question:} What was Warsaw's population in 1901?\\
\textbf{Answer:}\\
\textbf{Reference Answer:} 711,988
\end{quespromptbox}

\begin{quespromptbox}{SVAMP}
\textit{Answer the following question concisely in one sentence.}\\
\textbf{Context:} A grocery store had 72 bottles of regular soda, 32 bottles of diet soda and 78 apples.\\
\textbf{Question:} How many more bottles than apple did they have?\\
\textbf{Answer:}\\
\textbf{Reference Answer:} 26
\end{quespromptbox}

\begin{quespromptbox}{TriviaQA}
\textit{Answer the following question concisely in one sentence.}\\
\textbf{Question:} Traitor's Gate is part of which building?\\
\textbf{Answer:}\\
\textbf{Reference Answer:} tower of london
\end{quespromptbox}

\subsection{Datasets in Long-form Experiments}
We select DeepSeek-V3.1 and Gemini-3-Flash as representative SOTA LLMs at the time of writing. Due to the absence of fine-grained, long-form hallucination evaluation datasets for these two models, we construct four new datasets based on entities from FActScore \cite{min2023factscore} and PopQA \cite{mallen2023not}.

\textbf{FActScore:} This dataset is widely used for evaluating the factuality of biographies generated by LLMs, with entities sourced from Wikipedia. We select entities from the publicly released ``unlabelled'' portion and prompt each model with the query ``Tell me a bio of  wiki\_entity:[subject]'' to collect long-form generations.

\textbf{PopQA:} This dataset contains Wikipedia entries spanning 16 subject domains. Although PopQA is not originally designed for long-text generation, it includes long-tail entities with low-popularity, which present substantial challenges for offline-deployed LLMs. We filter for long-tail entities and prompt each model with the query ``Provide me with a paragraph detailing some facts related to [subject]'' to collect model-generated content.

To ensure quality, we further filter entities based on the informational completeness of their corresponding Wikipedia pages, removing those with page lengths shorter than 2000 tokens. This step ensures that each retained Wikipedia page contains sufficient reference material for subsequent annotation.

\textbf{Annotation Process:} For each model-dataset pair (four pairs in total), we employ a two-stage cross-validation annotation process. First, we generate factual claims for each sampled entity, following the processing pipeline detailed in Appendix~\ref{appendix:long-form generation}. Each claim that extracted from the greedily decoded response is then evaluated for factual accuracy according to the following criteria:

\begin{itemize}
    \item \textbf{True}: The claim is factually accurate and \textbf{can be verified by the corresponding Wikipedia pages.}
    \item \textbf{False}: The claim contains factual errors or contradicts information found on the corresponding Wikipedia page.
\end{itemize}
To enhance annotation efficiency and consistency, we utilize two powerful reasoning models, GPT-5 and DeepSeek-R1, for automated labeling. A claim is automatically assigned a label only if both models produce identical classifications; otherwise, the claim is reviewed by human annotators for final decision. This hybrid annotation framework ensures high accuracy while enabling large-scale processing. In total, \textbf{7,407} generated claims have been annotated through this process.

We find that even such powerful LLMs exhibit significant hallucination rates, with 28\% for DeepSeek-V3.1 and 25\% for Gemini-3-Flash.
Although hallucinations, like software vulnerabilities, cannot be completely eliminated at the current stage, effective detection mechanisms can mitigate associated risks and enhance the reliability of LLMs. Given that hallucinations may lead to serious real-world consequences, reliable uncertainty quantification is crucial for ensuring the responsible deployment of LLM-based systems, which highlights the important value of our research.

Below is the example of FactScore with DeepSeek-V3.1.

\begin{longformbox}{FActScore}{DeepSeek-V3.1}
    \boxsection{Prompt:} \\
    Tell me a paragraph bio of wiki\_entity: \href{https://en.wikipedia.org/wiki/Danny_Faure}{Danny Faure}
    
    \vspace{2mm} \hrule \vspace{2mm}

    \boxsection{Greedily Decoded Generation:} \\
    Danny Faure is a Seychellois politician who served as the President of Seychelles from 2016 to 2020. He assumed office following the resignation of his predecessor, James Michel, and became the fourth president of the island nation. Faure, a member of the United Seychelles party, had previously held key ministerial positions, including Vice President and Minister of Finance. During his presidency, he focused on economic reforms, environmental conservation, and social development. After leaving office, he was appointed as the Special Envoy for the African Union on maritime security in the Indian Ocean. Faure is recognized for his contributions to Seychelles' stability and sustainable development.
    
    \vspace{2mm} \hrule \vspace{2mm}

    \boxsection{Extracted Atomic Claims and Annotations:}
\begin{enumerate}[itemsep=0pt, topsep=2pt, parsep=0pt, partopsep=0pt, leftmargin=1.5em]
    \item[\textcolor{green!70!black}{\textsf{[T]}}] Danny Faure is a Seychellois politician.
    \item[\textcolor{green!70!black}{\textsf{[T]}}] Danny Faure served as the President of Seychelles from 2016 to 2020.
    \item[\textcolor{green!70!black}{\textsf{[T]}}] Danny Faure assumed office following the resignation of his predecessor, James Michel.
    \item[\textcolor{green!70!black}{\textsf{[T]}}] Danny Faure became the fourth president of Seychelles.
    \item[\textcolor{green!70!black}{\textsf{[T]}}] Danny Faure is a member of the United Seychelles party.
    \item[\textcolor{green!70!black}{\textsf{[T]}}] Danny Faure previously held the position of Vice President of Seychelles.
    \item[\textcolor{red}{\textsf{[F]}}] Danny Faure previously held the position of Minister of Environment in Seychelles.\\
    \textit{Annotation: Danny Faure previously held the position of Minister of Finance in Seychelles.}
    \item[\textcolor{red}{\textsf{[F]}}] During his presidency, Danny Faure focused on economic reforms.\\
    \textit{Annotation: economic reforms were clearly recorded only during his tenure as finance minister in Wikipedia.}

    \item[\textcolor{green!70!black}{\textsf{[T]}}] During his presidency, Danny Faure focused on environmental conservation.

    \item[\textcolor{red}{\textsf{[F]}}] During his presidency, Danny Faure focused on social development.\\
    \textit{Annotation: there is a lack of direct evidence to summarize the presidential term as "focusing on social development"}

    \item[\textcolor{red}{\textsf{[F]}}] Danny Faure was appointed as the Special Envoy for the African Union on maritime security in the Indian Ocean.\\
    \textit{Annotation: there is no such position.}

    \item[\textcolor{green!70!black}{\textsf{[T]}}] Danny Faure is recognized for his contributions to Seychelles' stability.

    \item[\textcolor{green!70!black}{\textsf{[T]}}] Danny Faure is recognized for his contributions to Seychelles' sustainable development.
    
\end{enumerate}

    \vspace{2mm} \hrule \vspace{2mm}
      
    \boxsection{Hallucination Analysis:}\\
    This serves as a typical example of how LLM long-form output interleaves true and false claims. While the model accurately describes most of the subject's historical political career of Danny Faure, it hallucinates a \textcolor{red}{nonexistent AU Special Envoy role} and exaggerates attributions that are not supported by any authoritative references.
\end{longformbox}

\section{Baselines Details}\label{appendix:Baselines}
This section aims to comprehensively elaborate on the benchmark methods employed in our experiments. We will conduct an in-depth analysis of the specific details of each method, including its prompts and implementation approaches. By providing detailed descriptions of these benchmark methods, we strive to ensure the reproducibility and transparency of the experimental setup.

\subsection{Baselines in Short-form Experiments}
\textbf{Length-normalized Predictive Entropy (LN-PE) \cite{malinin2021uncertaintyestimationautoregressivestructured}.}
To calculate prediction entropy, we must obtain the probabilities that LLMs assign to generated token sequences.
The probability of an entire sequence $s$ given context $x$ equals the product of probabilities for each token conditioned on previous tokens, with its log probability expressed as $\log P(s|x) = \sum_i \log P(s_i|s_{<i}, x)$. $s_i$ is the $i$-th output token and $s_{<i}$ denotes all preceding tokens. 
Due to the conditional independence of token probabilities, longer sequences inherently have lower joint likelihood. The joint likelihood of a sequence decreases exponentially with length $L$, while its negative log probability increases linearly with $L$, causing longer sentences to contribute disproportionately to entropy.
Length-normalized prediction entropy addresses this bias by normalizing the log probability by sequence length, using the arithmetic mean $\frac{1}{L} \sum_{i=1}^L \log P(s_i|s_{<i}, x)$.
This normalization effectively assumes that the uncertainty of the generated result is independent of sequence length.

\textbf{P(True) \cite{kadavath2022language}.}
This method first prompts LLMs to generate multiple distinct answers, then presents this answer list alongside the greedy decoding response and a binary question: ``Is this answer (a) correct or (b) incorrect?". 
The uncertainty score is determined by taking the negative of the probability that the LLMs answer "(a)" to this multiple-choice question.
Following \cite{farquhar2024detecting}, we enhance P(True) through few-shot prompting by incorporating ten randomly selected training examples formatted according to the described protocol and annotated with their true labels. 
This strategy represents a form of supervised in-context learning that leverages partial reference answer without necessitating model retraining.

\textbf{SelfCheckGPT (SC) \cite{manakul2023selfcheckgpt}.} 
SelfCheckGPT (SC) is the first zero-resource hallucination detection solution that can be applied to black-box systems. SC-Prompt represents its highest-performing variant, which prompts the LLM to assess the semantic consistency between the target sentence and a set of randomly generated samples, thereby determining the presence of hallucinations. The specific procedure is as follows: given a sentence to be evaluated $r_i$ and $N$ randomly generated samples $S^n$ corresponding to the same query, a fixed prompt template is used to instruct the LLM to evaluate their semantic consistency. The LLM's output is then converted into a numerical score $x$ in accordance with predefined rules. Finally, the inconsistency score for the sentence is obtained by averaging these individual scores. A score approaching 1.0 indicates a higher likelihood that the sentence contains hallucinations. The calculation formula is:
\begin{equation}
    {SC}_{prompt}=\frac{1}{N} \sum_{n=1}^{N} x^{n},
\end{equation}
where \(x_{i}^{n}\) is the mapping score corresponding to the \(n\)-th sample. In our experiment, we used the optimal variant ${SC}_{prompt}$ as the baseline.

\textbf{Embedding Regression (ER) \cite{farquhar2024detecting}.}
Embedding Regression represents a typical supervised learning approach. Inspired by Kadavath et al. \cite{kadavath2022language}, who developed predictors by fine-tuning proprietary language models on annotated QA datasets to assess whether target LLMs could correctly answer specific questions, \citep{farquhar2024detecting} implemented a more efficient alternative. This approach directly extracts the final hidden layer states from LLMs and trains an Embedding Regression classifier to achieve equivalent predictive functionality without requiring model fine-tuning or ground-truth answers.

\textbf{Semantic Entropy (SE) \cite{farquhar2024detecting}.} 
Semantic entropy aims to evaluate the uncertainty of LLMs regarding the meaning of their generated sequences. 
It first calculates the sum of probabilities for all token sequences that can be considered as expressing the same meaning. Given context $x$, for each semantic equivalence class $c$, its probability $P(c|x)$ is estimated through semantic clustering of generated sequences $s$:
\begin{equation}
P(c \mid x)=\sum_{s \in c} P(s \mid x)=\sum_{s \in c} \prod_{i} P\left(s_{i} \mid s_{<i}, x\right).
\end{equation}
Semantic entropy (SE) is then estimated as the Shannon entropy of the meaning distribution:
\begin{equation}
SE(x) = -\sum_{c \in C} P(c|x) \log P(c|x).
\end{equation}

In practice, it is not possible to calculate $\sum_C p(C \mid x) \log p(C \mid x)$ because of the intractable number of semantic clusters. Instead, discrete semantic entropy (DSE) uses a Rao-Blackwellized Monte Carlo estimator

\begin{equation}
    \mathrm{DSE}(x) \approx -\sum_{i=1}^{M} p'(C_i \mid x) \log p'(C_i \mid x),
\end{equation}

where $C_i$ are $M$ clusters extracted from the $N$ generations and $p'(C_i \mid x) = \frac{p(C_i \mid x)}{\sum_k p(C_k \mid x)}$, which we refer to as $p(C_i \mid x)$ in the following for simplicity. DSE can be extended to cases where token likelihoods are not available by approximating $p(C_i \mid x)$ with the fraction of generated texts in each cluster, $p(C_i \mid x) \approx \sum_{i=1}^{N} \mathbb{I}(s_i \in C_i) / N$.

\textbf{Kernel Language Entropy (KLE)\cite{nikitin2024kernel}.} 
To address the limitations of traditional semantic entropy, which only rely on semantic equivalence relations and fail to capture fine-grained semantic similarities, Nikitin et al. proposed Kernel Language Entropy (KLE). Its encodes the semantic similarities between texts generated by LLMs through a unit trace positive semidefinite kernel $K_{\text{sem}}$, and then use the von Neumann Entropy (VNE) to quantify the uncertainty of the semantic space represented by this kernel. For a unit trace positive definite matrix \( A \in \mathbb{R}^{n \times n} \), its von Neumann Entropy, denoted as \(\text{VNE}(A)\), is defined in the form of matrix trace operation. On this basis, the KLE is defined as the VNE of the semantic kernel:
\begin{align}
&\text{VNE}(A) = -\text{Tr}\left[A \log A\right], \\
& \text{KLE}(x) = \text{VNE}(K_{\text{sem}}).
\end{align}
In our experiment, we used the optimal variant $\text{KLE}_{HEAT}$ as the baseline, which is a heat kernel over constructed semantic graph.

\textbf{Semantic Graph Density (SGD)\cite{LiSYJCCR25}.} 
\citep{LiSYJCCR25} proposed Semantic Graph Density (SGD), which quantifies semantic consistency using graph density and adjusts edge contributions by incorporating answer probabilities. For \(N\) white-box sampled answers \(\{y^{(i)}\}_{i=1}^N\), pairwise semantic similarities \(s_{ij}\) and length-normalized probabilities \(P(y^{(i)}|x)\) are first computed to construct the semantic graph. SGD is defined as the negative value of semantic graph density, and the best-performing variant is version $\text{SGD}_{s+P}$, which optimizes edge contribution through a probability fusion strategy:
\begin{align}
&\text{SGD}_{s+P}(x) = -\sum_{i<j}s_{ij} \cdot \mu(i,j), \\
&\mu(i,j) = \theta \cdot \frac{1}{N(N-1)/2} + (1-\theta) \cdot \frac{P(y^{(i)}|x)P(y^{(j)}|x)}{\sum_{k<l}P(y^{(k)}|x)P(y^{(l)}|x)}.
\end{align}
Where \(\mu(i,j)\) is the edge contribution weight fusing prior uniform distribution and answer joint probability, and \(\theta\) is a balancing hyperparameter. In our experiment, we used the optimal variant $\text{SGD}_{s+P}$ as the baseline.

\subsection{Baselines in Long-form Experiments} 
\textbf{Discrete Semantic Entropy (DSE) variant \cite{farquhar2024detecting}.} 
For long-form generation, discrete semantic entropy operates through three key steps: 
(1) decomposing the text into atomic claims, 
(2) generating multiple possible questions that could trigger each claim in reverse, and (3) querying the original LLMs to produce new responses for each question, while including the original claim as a candidate. 
The final uncertainty estimate for each claim is derived by averaging the semantic entropy values across all associated questions. 
In our long-form experiments, we implement discrete semantic entropy according to the original paper's best practices. 
Applying discrete semantic entropy to long-form generation introduces additional assumptions and complexities, and its computational cost increases with higher sampling.
In contrast, SeSE achieves better performance with a more concise principle and lower cost.

\textbf{Verbalized Uncertainty (VU) \cite{mohri2024language}.} 
Verbalized Uncertainty prompts LLMs to directly express their confidence in a claim through natural language and maps confidence expressions in the LLMs' output (e.g., ``very confident", ``100\%", etc.) to numerical values. Uncertainty is quantified as the negative value of ther confidence score. Here, we mainly consider two variants:
\begin{itemize}
    \item \textbf{Post-hoc Verbalized Uncertainty (PH-VU):} This method elicits the verbalized confidence in a post-hoc manner after the entire claim set $C$ has been decomposed from generations. Specifically, we prompt an LLM to express its confidence about each claim $c \in C$ given multiple options such as ``Unlikely (40\%)'', ``Even chance (50\%)'' etc. The specific prompt that we adapted from \cite{mohri2024language} is as following:\\
\textit{You are provided with some possible information about a Wikipedia entity. Assess the likelihood that this information is correct and describe it using one of the following expressions: Certainly false (0\%), Very unlikely (20\%), Unlikely (40\%), Even chance (50\%), Possibly true (60\%), Likely (80\%), Certainly true (100\%).\\ 
Just provide your confidence expression, do not output other text or explanations. \\
The entity is: \{\textcolor{blue}{entity}\} \\
The possible information is: \{\textcolor{blue}{claim}\} \\
Output:}
\item \textbf{In-line Verbalized Uncertainty (IL-VU):} In-line verbalized uncertainty (IL-VU) directly elicits the verbalized confidence about each claim $c$ in an in-line manner right after it is decomposed from the generations. Thus, we prompt LLMs with a long-form generation and instructions to give all the claims with corresponding confidence scores. The specific prompt that we adapted from \cite{mohri2024language} is as following:\\
\textit{Please deconstruct the input paragraph into the smallest possible standalone self-contained facts without semantic repetition with corresponding confidence score.\\ 
The confidence score should represent your confidence in the claim, where a 1 is obvious facts and results like $1+1=2$. A 0 represents claims obviously incorrect or difficult for anyone to understand, such as ``The Earth is the center of the universe" or ``The exact population of a certain ordinary town". \\
You must return the output as a jsonl, where each line is claim:\{[claim],[confidence score]\}.\\
The input is: \{\textcolor{blue}{long-form generation}\}\\
Output:
}
\end{itemize}

\textbf{P(True) variant.}
P(True) estimates the uncertainty of a claim by prompting LLMs to answer whether a claim is true or false, using the negative probability of the claim being true as the uncertainty score. 
Since probabilities cannot be obtained directly in long-form experiments, we improve this method by prompting the model to answer 10 times and estimating the probability by calculating the frequency of ``true" responses. 
The specific prompt that we adapted from \cite{kadavath2022language} is as following:
\begin{itemize}[leftmargin=\parindent, labelindent=0pt]
\item[]\textit{Is the claim true or false? Answer with only True or False. \\
Claim: \{\textcolor{blue}{claim}\}\\
Output:
}
\end{itemize}

\textbf{SC variant.}
This methods utilizes the consistency score of one claim across different samples, and the consistency score is determined by prompting the same LLM. The prompt used to evaluate consistency is described detailed in Appendix~\ref{appendix:long-form generation}.

\end{document}